\documentclass[sigconf]{acmart}

\pdfpageattr {/Group << /S /Transparency /I true /CS /DeviceRGB>>}

\usepackage{balance}
\usepackage{enumitem}
\usepackage{multirow}
\usepackage{booktabs}
\usepackage{caption}
\usepackage{array}
\usepackage{makecell}
\usepackage{color}
\usepackage{xcolor}

\usepackage{subcaption}
\usepackage{algorithm}
\usepackage{algpseudocode}
\usepackage{threeparttable}
\usepackage{pifont}
\usepackage{pdfpages}
\usepackage{pdfcomment}
\usepackage{tikz}
\usepackage{amsmath}
\usepackage{pifont}
\usepackage{float}
\usepackage{bm}
\usepackage{listings} 
\usepackage{xcolor} 
\usepackage{cuted}
\usepackage{capt-of}

\definecolor{deepgreen}{rgb}{0,0.5,0} 

\lstdefinestyle{lfonts}{
    basicstyle = \footnotesize\ttfamily,
    stringstyle = \color{deepgreen},
    keywordstyle = \color{blue!60!black}\bfseries,
    commentstyle = \color{gray}\scshape,
}
\lstdefinestyle{lnumbers}{
    numbers = left,
    numberstyle = \tiny,
    numbersep = 1em,
    firstnumber = 1,
    stepnumber = 1,
}
\lstdefinestyle{llayout}{
    breaklines = true,
    tabsize = 2,
    columns = flexible,
}
\lstdefinestyle{lgeometry}{
    xleftmargin = 20pt,
    xrightmargin = 0pt,
    frame = tb,
    framesep = \fboxsep,
    framexleftmargin = 20pt,
}
\lstdefinestyle{lgeneral}{
    style = lfonts,
    style = lnumbers ,
    style = llayout,
    style = lgeometry,
}
\lstdefinestyle{python}{
    language = {Python},
    style = lgeneral,
}

\definecolor{deepred}{rgb}{0.6,0,0} 
\definecolor{darkblue}{rgb}{0,0,0.5} 

\lstdefinestyle{xmlfonts}{
    basicstyle = \footnotesize\ttfamily,
    stringstyle = \color{deepred}, 
    tagstyle = \color{darkblue}\bfseries, 
    commentstyle = \color{gray}\itshape, 
}

\lstdefinelanguage{XML}{
    morestring=[b]",
    morecomment=[s]{<?}{?>},
    morecomment=[s]{<!--}{-->},
    tag=[s],
    otherkeywords={/>},
}

\lstdefinestyle{xmlnumbers}{
    numbers = left,
    numberstyle = \tiny,
    numbersep = 1em,
    firstnumber = 1,
    stepnumber = 1,
}
\lstdefinestyle{xmllayout}{
    breaklines = true,
    tabsize = 2,
    columns = flexible,
}
\lstdefinestyle{xmlgeometry}{
    xleftmargin = 20pt,
    xrightmargin = 0pt,
    frame = tb,
    framesep = \fboxsep,
    framexleftmargin = 20pt,
}
\lstdefinestyle{xmlgeneral}{
    style = xmlfonts,
    style = xmlnumbers,
    style = xmllayout,
    style = xmlgeometry,
    language = XML,
}
\lstdefinestyle{xml}{
    style = xmlgeneral,
}

\newcommand{\cmark}{\textcolor{my_green}{\ding{51}}} 
\newcommand{\xmark}{\textcolor{my_red}{\ding{55}}} 
\definecolor{my_green}{RGB}{51,102,0}
\definecolor{my_red}{RGB}{204, 0, 0}
\AtBeginDocument{%
  \providecommand\BibTeX{{%
    \normalfont B\kern-0.5em{\scshape i\kern-0.25em b}\kern-0.8em\TeX}}}

\copyrightyear{2026}
\acmYear{2026}
\setcopyright{cc}
\setcctype{by}
\acmConference[WSDM '26]{Proceedings of the Nineteenth ACM International Conference on Web Search and Data Mining}{February 22--26, 2026}{Boise, ID, USA}
\acmBooktitle{Proceedings of the Nineteenth ACM International Conference on Web Search and Data Mining (WSDM '26), February 22--26, 2026, Boise, ID, USA}
\acmPrice{}
\acmDOI{10.1145/3773966.3777972}
\acmISBN{979-8-4007-2292-9/2026/02}

\makeatletter
\gdef\@copyrightpermission{
  \begin{minipage}{0.3\columnwidth}
   \href{https://creativecommons.org/licenses/by/4.0/}{\includegraphics[width=0.90\textwidth]{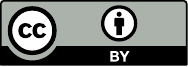}}
  \end{minipage}\hfill
  \begin{minipage}{0.7\columnwidth}
   \href{https://creativecommons.org/licenses/by/4.0/}{This work is licensed under a Creative Commons Attribution International 4.0 License.}
  \end{minipage}
  \vspace{5pt}
}
\makeatother

\begin{document}

\title{STORM: A Spatio-Temporal Factor Model Based on Dual Vector Quantized Variational Autoencoders for Financial Trading}

\renewcommand{\shorttitle}{A Spatio-Temporal Factor Model Based on Dual Vector Quantized Variational Autoencoders for Financial Trading}
\author{Yilei Zhao}
\authornote{These authors contributed equally to this work.}
\affiliation{%
  \institution{Nanyang Technological University}
  \country{Singapore}
}
\email{YILEI002@e.ntu.edu.sg}

\author{Wentao Zhang}
\authornotemark[1]
\affiliation{%
  \institution{Nanyang Technological University}
  \country{Singapore}
}
\email{zhangwent963@gmail.com}

\author{Tingran Yang}
\affiliation{%
  \institution{Zhejiang University}
  \city{Hangzhou}
  \state{Zhejiang}
  \country{China}
}
\email{3220102076@zju.edu.cn}

\author{Yong Jiang}
\affiliation{%
  \institution{Alibaba Group}
  \city{Hangzhou}
  \state{Zhejiang}
  \country{China}
}
\email{jiangyong.ml@gmail.com}

\author{Fei Huang}
\affiliation{%
  \institution{Alibaba Group}
  \city{Hangzhou}
  \state{Zhejiang}
  \country{China}
}
\email{f.huang@alibaba-inc.com}

\author{Wei Yang Bryan Lim}
\authornote{Corresponding author.} 
\affiliation{%
  \institution{Nanyang Technological University}
  \country{Singapore}
}
\email{Bryan.limwy@ntu.edu.sg}

\renewcommand{\shortauthors}{Yilei Zhao et al.}

\begin{abstract}
In financial trading, factor models are widely used to price assets and capture excess returns from mispricing. Recently, we have witnessed the rise of variational autoencoder-based latent factor models, which learn latent factors self-adaptively. While these models focus on modeling overall market conditions, they often fail to effectively capture the temporal patterns of individual stocks. Additionally, representing multiple factors as single values simplifies the model but limits its ability to capture complex relationships and dependencies. As a result, the learned factors are of low quality and lack diversity, reducing their effectiveness and robustness across different trading periods.
To address these issues, we propose a \textbf{S}patio-\textbf{T}emporal fact\textbf{OR} \textbf{M}odel based on dual vector quantized variational autoencoders, named \texttt{STORM}, which extracts features of stocks from temporal and spatial perspectives, then fuses and aligns these features at the fine-grained and semantic level, and represents the factors as multi-dimensional embeddings. The discrete codebooks cluster similar factor embeddings, ensuring orthogonality and diversity, which helps distinguish between different factors and enables factor selection in financial trading. To show the performance of the proposed factor model, we apply it to two downstream experiments: portfolio management on two stock datasets and individual trading tasks on six specific stocks. The extensive experiments demonstrate \texttt{STORM}'s flexibility in adapting to downstream tasks and superior performance over baseline models. \textbf{Code is available in PyTorch\footnote{https://github.com/DVampire/Storm}}.

\end{abstract}




\maketitle

\section{Introduction}

In financial trading, factor models are fundamental tools for asset pricing and are widely used to predict asset returns. These models enhance pricing accuracy and risk management by identifying a set of key factors that explain excess returns. The well-known Fama-French three-factor model \cite{fama1992cross} extends the Capital Asset Pricing Model \cite{sharpe1964capital} with additional factors to capture market risk premiums. As financial theories evolve, researchers continue to discover new factors, broadening our understanding of market dynamics.

\begin{figure}[t]
\setlength{\abovecaptionskip}{0.05cm}
\centering
\includegraphics[width=0.45\textwidth]{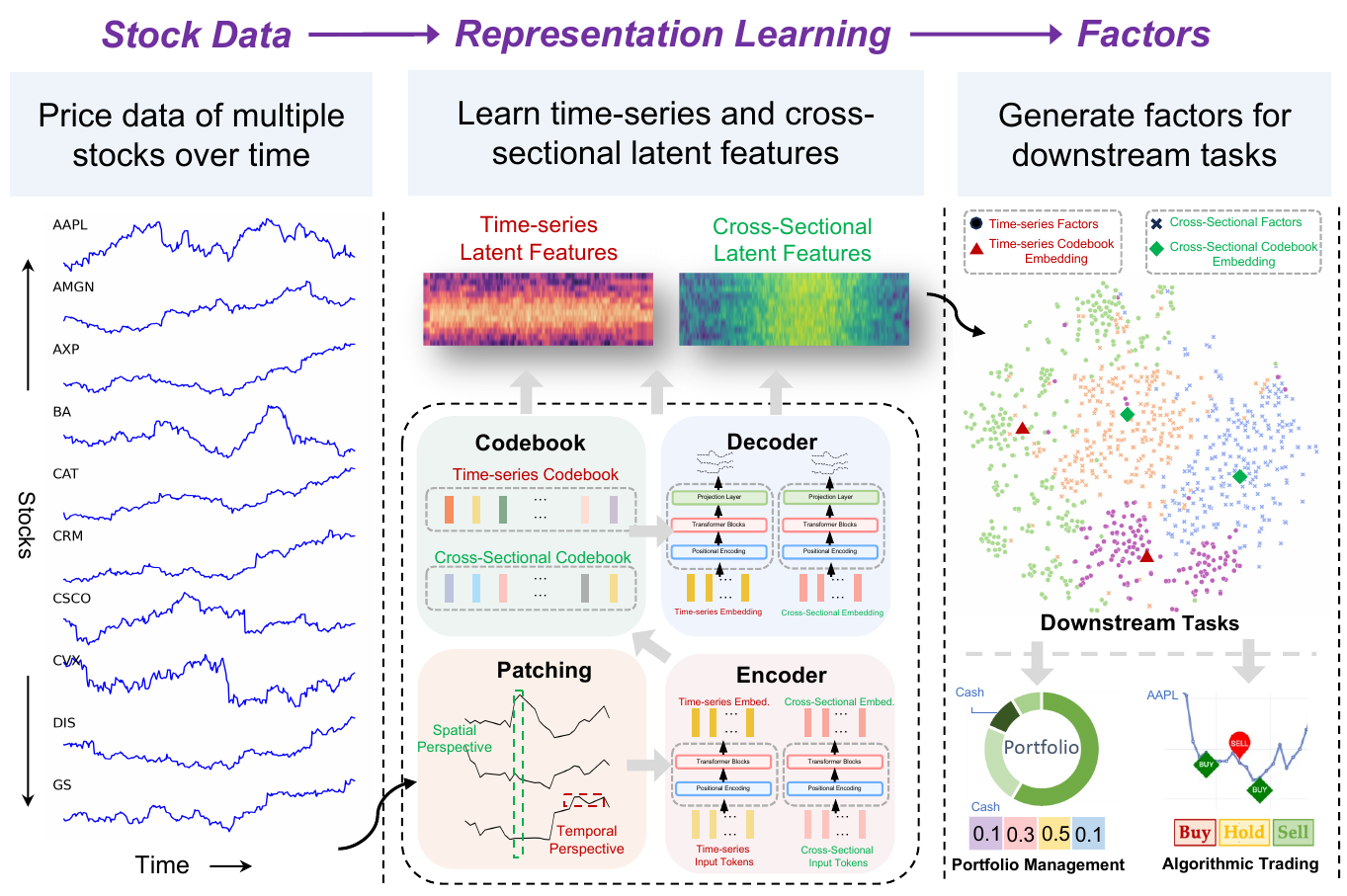}
\caption{Overall architecture of \texttt{STORM}}
\label{fig:overview}
\end{figure}

Recently, we have witnessed the rise of latent factor models \cite{gu2021autoencoder,duan2022factorvae,wei2022factor,wei2023hirevae}, connecting the factor model with the generative model, the variational autoencoder (VAE). These VAE-based models describe high-dimensional data (prices) to low-dimensional representations (factors), and learn factors self-adaptively. Although latent factor models have demonstrated substantial success in financial trading tasks, they still face several significant issues:
\begin{itemize}[left=0.3em]
    \item \textbf{CH1: Limited Reflection of Market Complexity.} Latent factor models represent factors as single values are inherently constrained by their insufficient capacity to capture the intricate complexity and nonlinearity of financial data, rendering them vulnerable to noise and non-stationarity, which compromises their predictive accuracy and stability.
    \item \textbf{CH2: Factor Inefficiency.} VAE-learned factors suffer from three inefficiencies: i) focusing mainly on cross-sectional factors while neglecting temporal information, ii) allowing noise in continuous latent spaces to overshadow meaningful signals, and iii) lacking independence among factors, which leads to multicollinearity and weak adaptability to varying market conditions.
    \item \textbf{CH3: Lack of Factor Selection.} Existing latent factor models primarily focus on generating factors without adequately differentiating between them. Furthermore, they neglect the crucial process of factor selection, which is essential for identifying impactful factors, thereby limiting the model's overall effectiveness and precision.
\end{itemize}

In order to address the challenges, we propose a \textbf{S}patio-\textbf{T}emporal fact\textbf{OR} \textbf{M}odel based on dual vector quantized variational autoencoders (VQ-VAE), named \texttt{STORM}, with the architecture shown in Figure \ref{fig:overview}. Unlike traditional scalar-valued financial factors with clear economic interpretations, \texttt{STORM} learns high-dimensional latent vectors that, while less interpretable capture complexity and nonlinearity inherent in financial martket (\textbf{CH1}).
Additionally, we develop a dual VQ-VAE architecture to capture cross-sectional and time-series features, considering both spatial and temporal perspectives\footnote{We treat cross-sectional as spatial and time-series as temporal. These terms will be used interchangeably unless specified.}. By integrating these features at both fine-grained and semantic levels, the model constructs more effective factors. Through diversity and orthogonality loss constraints, we ensure representation independence among learned representations (\textbf{CH2}). Furthermore, codebook embeddings act as cluster centers, serving as class tokens to categorize factor embeddings. This strategy provides clarity and transparency to the differentiation and selection process of factors (\textbf{CH3}).
Specifically, our contributions are four-fold:

\begin{itemize}[left=0em]
    \item We design a dual VQ-VAE architecture that systematically constructs cross-sectional and time-series factors by capturing both spatial and temporal dependencies.
    \item We leverage vector quantization techniques to improve factor embedding representation by reducing noise, enhancing diversity, and ensuring orthogonality. This approach significantly strengthens the factors' predictive power and a more transparent factor selection process.
    \item We develop \texttt{STORM} as a versatile all-rounder capable of supporting three tasks simultaneously, whereas other methods tend to be specialized in specific tasks.
    \item Experiments on seven datasets from two U.S. stock markets, covering the stock future return prediction task (DJ30 and SP500), portfolio management (DJ30 and SP500), and algorithmic trading (five stocks), show that \texttt{STORM} outperforms all baselines across two prediction metrics and six standard financial metrics. 
\end{itemize}

\section{Related Work}

\subsection{Factor Model}
Factor model has attracted extensive research focus from the finance community. Financial experts have carefully selected a wide range of factors from macroeconomic \cite{sharpe1964capital}, market changes \cite{fama1992cross}, individual stock fundamentals \cite{greenblatt2010little}, and other perspectives to price assets in an attempt to capture excess returns from mispricing.
Recently, latent factor models have emerged that improve investment returns by studying latent factors self-adaptively. Specifically, \cite{gu2021autoencoder} learns latent factors depend on asset characteristics, \cite{duan2022factorvae} extracts cross-sectional factors by predicting future returns, and \cite{wei2023hirevae} learns market-augmented latent factors. However, these VAE-based latent factor models may suffer from a lack of robustness in representing low signal-to-noise ratio markets, leading to reduced reconstruction and generation capabilities. Furthermore, their effectiveness is constrained by an insufficient focus on the temporal perspective.

\subsection{Financial Trading}
Financial trading has attracted significant attention from both the finance and AI communities. Quantitative traders leverage mathematical models and algorithms to automatically identify trading opportunities \cite{sun2024trademaster}. Within the FinTech domain, key research areas encompass portfolio management (PM), algorithmic trading (AT), order execution \cite{fang2021universal}, market making \cite{spooner2020robust}, etc. In this work, we focus on PM and AT, as they are fundamental to developing intelligent financial decision-making systems.

PM focuses on optimizing wealth allocation across assets \cite{li2014online}. Its development has evolved from simple rule-based methods \cite{poterba1988mean} to prediction-based approaches using machine learning \cite{ke2017lightgbm} and deep learning \cite{qin2017dual,zhao2022stock} for predicting returns or price movements. Despite recent integration of reinforcement learning (RL) in PM \cite{zhang2024reinforcement}, the latent factor models still predominantly rely on deep learning for market modeling and precise predictions.

AT, on the other hand, involves executing trades with algorithm-generated signals. Like PM, AT has seen a transition from rule-based and supervised learning techniques \cite{yang2020qlib} to RL-driven approaches. In particular, RL-based methods \cite{sun2022deepscalper} have gained prominence due to their ability to navigate complex sequential decision-making problems, further enhancing trade execution efficiency and adaptability in dynamic market environments.


\subsection{Vector Quantized Variational Autoencoder}
Vector Quantized Variational Autoencoder (VQ-VAE) has emerged as a fundamental framework in deep learning research, establishing methodological significance across core domains: representation learning, generative modeling, and time series analysis. Initially introduced by \citeauthor{van2017neural}\cite{van2017neural}, VQ-VAE has showcased its capabilities in processing high-dimensional data across diverse modalities, including images \cite{razavi2019generating}, audio \cite{zhang2024codebook}, and video \cite{yan2021videogpt}, with a strong emphasis on learning discrete latent representations. Particularly noteworthy is the model's demonstrated proficiency in capturing temporal dependencies, where recent studies like \cite{talukder2024totem} have validated its superior performance in modeling sequential patterns compared to conventional continuous latent space approaches, thereby advancing the development of robust generative architectures specifically optimized for temporal dynamics analysis.
\section{Preliminaries}

\subsection{Factor Model}

The factor model (e.g., Arbitrage Pricing Theory \cite{ross1976arbitrage}) is defined to study the relationship between expected asset returns and factor exposures, focusing on the cross-sectional differences in expected asset returns. 
\begin{equation}
    \mathbb{E}[R_{i}^{e}] = \alpha_{i} + \beta_{i} \lambda \text{.}
\end{equation}
Here, $\mathbb{E}[\cdot]$ denotes the expectation operator, and $\mathbb{E}[R_{i}^{e}]$ represents the expected excess return of asset $i$. $\beta_{i}$ is factor exposure (or factor loading) of asset $i$, and $\lambda$ is factor expected return (or factor risk premium). Lastly, $\alpha_{i}$ captures pricing error, reflecting any deviations from the model’s predicted returns.

According to \cite{duan2022factorvae}, the factor model is formulated from a spatial perspective, primarily examining cross-sectional variations in expected returns across different assets, rather than capturing the temporal evolution of individual asset returns. Therefore, we expand it from the temporal perspective and introduce the general functional form of the latent factor model:
\begin{equation}
\label{equ: factor model}
    y_{i,t}=\alpha_{i,t}+\sum_{k=1}^K \boldsymbol{\beta}_{i,t}^k \mathbf{z}_{i,t}^k + \epsilon_{i,t} \text{,}
\end{equation}
where $y_{i,t} $ denotes the future return of stock $i$ at timestep $t$, $\mathbf{z}_{i,t}$ denotes $K$ factor returns, and $\epsilon_{i,t} $ is the noise. From the spatial, Equation \ref{equ: factor model} can be expressed as $\mathbf{y}_{i}=\boldsymbol{\alpha}_{i}+\sum_{k=1}^K \boldsymbol{\beta}_{i}^k \mathbf{z}_{i}^k + \boldsymbol{\epsilon}_{i}$, which we regard as the cross-sectional factor model. From the temporal perspective, Equation \ref{equ: factor model} can be expressed as $\mathbf{y}_{t}=\boldsymbol{\alpha}_{t}+\sum_{k=1}^K \boldsymbol{\beta}_{t}^k \mathbf{z}_{t}^k + \boldsymbol{\epsilon}_{t}$, which we regard as the time-series factor model. The cross-sectional factor model focuses on the differences in returns of different assets at the same point in time, while the time-series factor model focuses on the changes in returns of a single asset at different time points.

\subsection{Problem Formulation}
To assess the effectiveness and generating ability of factors, we predict and evaluate the stock future returns. To show the adaptability of the latent factor model with multiple downstream tasks, we employ two financial trading tasks, i.e., \textit{portfolio management} \cite{wang2021deeptrader, zhang2024reinforcement, ye2020reinforcement} and \textit{algorithmic trading} \cite{sun2022deepscalper, liu2020adaptive}, to showcase the model's investment capabilities.

\textbf{Observed Data}. We utilize the stock's historical price data $\mathbf{p}$ (i.e., open, high, low, and close) and technical indicators  $\mathbf{d}$ as observed variables $\mathbf{x} := [\mathbf{p}, \mathbf{d}] \in \mathbb{R}^{N\times W \times D}$, within a window size of $W$. The historical price data of stock $i$ at time $t$ is denoted as $\mathbf{p}_{i,t}:=[p_{i,t}^o,p_{i,t}^h,p_{i,t}^l,p_{i,t}^c] \in \mathbb{R}^{D_1}$. The technical indicators which are calculated from price and volume values, are denoted as $\mathbf{d}_{i,t} \in \mathbb{R}^{D_2}$. Each stock is represented by $D=D_1+D_2$ features per trading day, which we then use to predict the stock's future returns, defined as $\mathbf{y}_{i,t+1}:=\frac{p_{i,t+1}-p_{i,t}}{p_{i,t}}$.

\textbf{Downstream tasks}. These primarily encompass the two most widely studied areas in quantitative finance -- \textit{Portfolio management} and \textit{Algorithmic trading}:
\begin{itemize}[left=0em]
    \item \textit{Portfolio management} task aims to construct an optimal portfolio based on the predicted returns to test the profitability of the factor model. At each trading day $t$, the factor model outputs return predictions $\hat{\mathbf{y}}_{i,t+1}$ for each stock $i$, which are then used by a portfolio allocation strategy, e.g., \textit{TopK-Drop}, to generate the portfolio weight vector $\mathbf{w}_t = [w_t^0,w_t^1,\dots,w_t^{N-1}] \in \mathbb{R}^{N}$, where $w_{t}^{i}$ represents the proportion of capital invested in stock $i$. To ensure full capital deployment, the allocation must satisfy the budget constraint $\sum_{i=0}^{N-1} w_{t}^{i} = 1$.

    \item \textit{Algorithmic trading} task aims to execute buy, hold, and sell actions based on predicted asset states to balance returns and risks. We formulate it as a Markov Decision Process (MDP) under reinforcement learning scenario following \cite{chen2023mtrader,zhang2024finagent}. The MDP is constructed by a 5-tuple $(\mathit{S}, \mathcal{A}, \mathcal{T}, \mathcal{R}, \gamma)$, where $\mathit{S}$ and $\mathcal{A}$ are sets of states and actions respectively, $\mathcal{T}:\mathit{S} \times \mathcal{A} \times \mathit{S} \rightarrow [0, 1]$ is the state transition function, $\mathcal{R}: \mathit{S} \times \mathcal{A} \rightarrow \mathbb{R}$ is the reward function where $\mathbb{R}$ is a continuous set of possible rewards, and $\gamma \in [0, 1)$ is the discount factor. The latent factor embeddings $\mathbf{Z}$ generated by the factor model are integrated into the observation set that constitutes the state space $\mathcal{S}$, enabling the agent to incorporate predictive market signals. The goal is to learn a policy $\mathcal{\pi} : \mathit{S} \rightarrow \mathcal{A}$ that maximizes the expected discounted cumulative reward $\mathbb{E}\lbrack\sum^{\infty}_{t=0}\gamma^{t-1} r_{t}\rbrack$. At each trading day $t$ for specific stock $i$, the agent takes action $a_{i,t}:=\{a_{i,t}^b, a_{i,t}^h, a_{i,t}^s\} \in \mathcal{A}$ which means buy, hold and sell, according to the current environment $s_t \in \mathcal{S}$. 
\end{itemize}

\subsection{Vector Quantized Variational Autoencoder}

VQ-VAE\cite{van2017neural} is different from VAE \cite{kingma2013auto} as it quantizes the observed data into a discrete token sequence. The VQ-VAE contains three parts, encoder, decoder, and discrete codebook, denoted by $\phi_{\text{enc}}$ with parameter $\theta_{\text{enc}}$, $\phi_{\text{dec}}$ with parameter $\theta_{\text{dec}}$ and $\mathcal{C}=\{(k,e_k\in R^{H})\}_{k=1}^K$ with parameter $\{e_k \in R^{H}\}_{k=1}^K$ respectively. Among them, $K$ is the size of the codebook, $H$ is the dimension of the vector, and $e_k$ is the $k$-th vector of the latent embedding space. Given an input $x$, the encoder $\phi_{\text{enc}}$ firstly encodes it into a set of continuous feature vectors $Z = \phi_{\text{enc}}(x)$ and quantization operation $q(\cdot)$ passes each feature vector $z \in Z$ through the discretization bottleneck following by selecting the nearest neighbor embedding in the codebook $\mathcal{C}$ as its discrete code sequence $Q_z$.
\begin{equation}
    Q_z = q(z,\mathcal{C})=c, \text{where } c=\arg\min_{k \in [0, K-1]}||z-e_k||_2^2 \text{.}
\end{equation} Then, the quantized feature for $z$ denoted as $Z^q_z$, is obtained through $Z^q_z = e_{Q_z}$. The quantized vectors $Z^q$ is fed into the decoder $\theta_{\text{dec}}$ to reconstruct the original data $x'= \theta_{\text{dec}}(Z^q)$.

\begin{figure*}[thb]
\setlength{\abovecaptionskip}{0.03cm}
\centering
\includegraphics[width=0.85\linewidth]{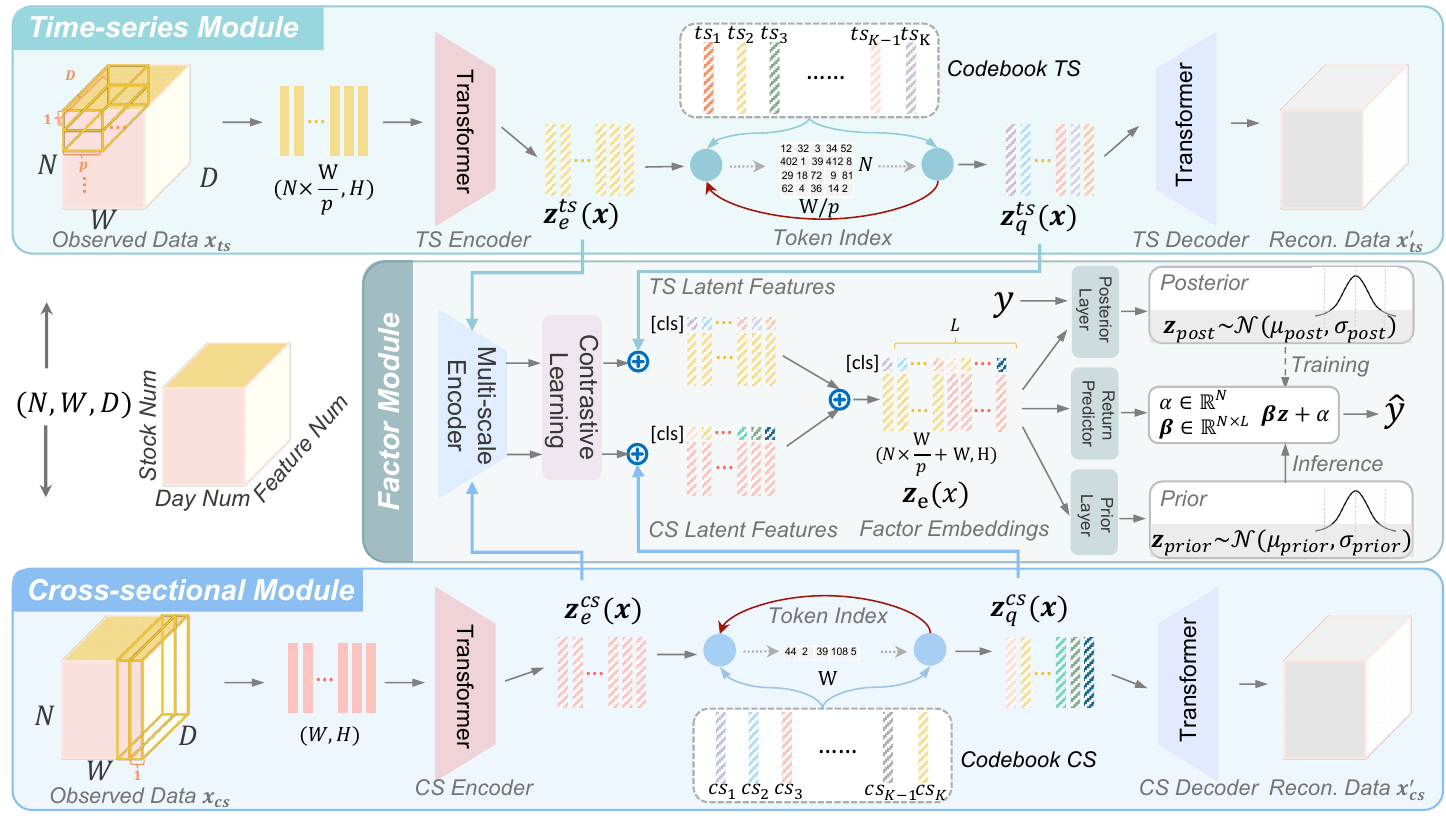}
\caption{The architecture of our proposed \texttt{STORM} model}
\label{fig: model}
\end{figure*}

\section{Method}
In this section, we represent our proposed latent factor model \texttt{STORM}, with the overall structure of the VQ-VAE depicted in Figure \ref{fig: model}. In general, \texttt{STORM} models data into time-series and cross-sectional features, deriving factors by integrating both temporal and spatial perspectives. Firstly, we introduce the dual VQ-VAE architecture and explain the extraction of features in two modules. Then the factor module will fuse and align features and generate factor embeddings. Finally, we demonstrate the effectiveness of \texttt{STORM} through comprehensive evaluations on two critical downstream tasks: portfolio management and algorithmic trading.

\subsection{Dual VQ-VAE Structure}
\label{sec: Dual Encoder-Decoder Structure}

The dual VQ-VAE architecture processes stock data from both spatial and temporal perspectives by \textit{patching} the data within each module. The \textit{encoder} then extracts time-series (TS) and cross-sectional (CS) features, which are subsequently clustered using a learnable \textit{codebook} for quantized representation. Finally, the \textit{decoder} reconstructs the data, effectively capturing complex patterns and enhancing model's forecasting performance.

\subsubsection{Patching}
In stock data analysis, it is essential to extract features from both temporal and spatial dimensions. Time-series features capture the dynamics of a stock over time, while cross-sectional features reveal the correlations among different stocks. To achieve this, we partition data into patches \cite{nie2022time} within the TS and CS modules. In the TS module, the observed data $\mathbf{x}$ is divided along the stock number dimension,  with each patch containing data for one stock over $p$ days (as illustrated in Figure~\ref{fig: model}). Similarly, in the CS module, the observed data $\mathbf{x}$ is divided along the time axis, with each patch containing features of all stocks for a single trading day. Combining information from multi-scale features allows the model to capture both temporal and spatial dependencies, enhancing the depth and accuracy of market insights.

\subsubsection{TS and CS Encoders}

Considering the Transformer's \cite{vaswani2017attention} remarkable ability to capture both local and global dependencies via self-attention and patching strategies, it is particularly well-suited for long-term time-series forecasting \cite{nie2022time}. Therefore, we employ stacked transformer blocks as the encoders in both the TS and CS modules to capture complex time-series and cross-sectional patterns. This architecture enhances the accuracy and robustness of feature extraction, ultimately improving forecasting performance.

While transformers are widely applicable across domains such as natural language processing (NLP) and computer vision (CV), \texttt{STORM}’s flexible encoder-decoder framework allows for the integration of alternative architectures, such as long short-term memory (LSTM) \cite{wang2021clvsa} and temporal convolutional networks (TCN) \cite{yu2015multi}. 

An encoder–decoder configuration comprising 4 and 2 blocks, respectively, each with 4 and 8 attention heads, remains relatively lightweight, avoiding excessive computational cost and a substantial increase in model complexity.

\subsubsection{Codebook Construction and Optimization}
The construction of the codebook involves a learnable embedding space designed to quantize the encoded features into discrete representations \cite{van2017neural}. 
During training, the continuous latent features $\mathbf{z}_e^{ts}(\mathbf{x})$ and $\mathbf{z}_e^{cs}(\mathbf{x})$ are quantized into discrete space by mapping each vector to its nearest codebook entry, $\mathbf{z}_q^{ts}(\mathbf{x})$ and $\mathbf{z}_q^{cs}(\mathbf{x})$, based on minimizing the Euclidean distance. This process translates the encoded features into discrete tokens, enabling the model to leverage a finite set of vectors to capture complex patterns. 

The VQ-VAE offers significant advantages in the latent factor model, primarily through its use of discrete codebook embeddings:
i) \textit{Discretization Benefits}: The discretization process inherent in VQ-VAE helps in clustering distinct and meaningful factors, providing orthogonality and diversity which improves upon the factors derived from traditional methods like FactorVAE. ii) \textit{Explicit Factor Selection}: The discrete token indices facilitate an explicit factor selection process, identifying the most relevant factors that influence stock returns and thus enhancing prediction accuracy. iii) \textit{Noise Reduction}: By filtering out irrelevant or redundant information during factor selection, VQ-VAE mitigates noise, contributing to improved model robustness and more reliable financial insights.

To encourage equal usage of the codebook vectors \cite{baevski2020wav2vec}, we incorporate a diversity loss to ensure balanced representations across the codebook. It is achieved by maximizing the entropy of the averaged softmax distribution over vectors for each codebook $\bar{p}_{g}$:
\begin{equation}
\mathcal{L}_{\text{div}}=\frac{1}{GK}\sum_{g=1}^{G}\sum_{k=1}^{K}\bar{p}_{g,k} \log{\bar{p}_{g,k}} \text{,}
\end{equation}where $G$ represents the number of codebooks, set 2 in \texttt{STORM}. The diversity constraint helps enhance representational capacity, leading to better coverage of the input space.

However, the lack of independence among factors can introduce multicollinearity, making the model prone to overfitting and reducing its robustness across different market conditions. To address this issue, we impose an orthogonality constraint on the codebook, as recent studies \cite{shin2023exploration} have shown that enforcing orthogonality allows discretized codes to maintain translation equivariance. The orthogonality loss is:
\begin{equation}
    \mathcal{L}_{\text{ortho}}= \frac{1}{K^2}\left|\left| \ell_2(\mathbf{e})^\top \ell_2(\mathbf{e}) - I_K \right|\right|_F^2 \text{,}
\end{equation}where $I_k \in \mathbb{R}^{K \times K}$ is the identity matrix, $\mathbf{e}$ is a surrogate for two codebook embeddings $\mathbf{ts}$ and $\mathbf{cs}$, $\ell_2(\mathbf{e})$ denotes $L_2$-normalized embeddings, and $\left|\left| \cdot \right|\right|_F$ is the Frobenius norm. The orthogonal constraint ensures that factors are independent of each other, allowing the effect of each factor on returns to be individually explained, which is crucial for asset pricing analysis.

\subsection{TS and CS Decoders}
The decoder is responsible for reconstructing the original data $\mathbf{x}$ from the quantized latent vectors $\mathbf{z}_q^{ts}(\mathbf{x})$ and $\mathbf{z}_q^{cs}(\mathbf{x})$. We utilize Transformer as decoders either, to generate constructed data $\mathbf{x}_{ts}'$ and $\mathbf{x}_{cs}'$, which aim to closely approximate the original input data.

Finally, the encoders, decoders, and codebooks are jointly optimized by minimizing the following loss objectives:
\begin{equation}
    \begin{aligned}
        \mathcal{L}_1 = & \lambda_{\text{ortho}}\mathcal{L}_{\text{ortho}} + \lambda_{\text{div}}\mathcal{L}_{\text{div}} + \left| \left|\mathbf{x}-\mathbf{x}'_{ts}\right|\right|^2_2+\left|\left| \mathbf{x}-\mathbf{x}'_{cs}\right|\right|^2_2 \\
        + & \left|\left|sg[\mathbf{z}_e^{ts}(\mathbf{x})]-\mathbf{z}_q^{ts}(\mathbf{x})\right|\right|^2_2 + \left| \left|sg[\mathbf{z}_q^{ts}(\mathbf{x})]-\mathbf{z}_e^{ts}(\mathbf{x})\right|\right|^2_2\\
        + & \left|\left|sg[\mathbf{z}_e^{cs}(\mathbf{x})]-\mathbf{z}_q^{cs}(\mathbf{x})\right|\right|^2_2 + \left|\left| sg[\mathbf{z}_q^{cs}(\mathbf{x})]-\mathbf{z}_e^{cs}(\mathbf{x})\right|\right|^2_2  \text{,}\\
    \end{aligned}
\end{equation}where $sg[\cdot]$ is a stop-gradient operator.

\subsection{Factor Module}
The factor module aims to integrate the TS and CS latent features through \textit{feature fusion and alignment}.
Afterward, it utilizes \textit{prior-posterior learning} to generate factor embeddings from latent features to predict stock future returns.

\subsubsection{Feature Fusion and Alignment}
As mentioned above we employ different patching methods to ensure that both time-series and cross-sectional features of the data were accurately captured. However, these methods result in two parallel processes that do not align at the feature level. To address the limitation, we employ a \textit{cross-attention mechanism} to enhance the interaction and fusion of features at a fine-grained level, improving the model's dynamic understanding of the input data. Additionally, \textit{contrastive learning} is used at the semantic level to enhance feature representation by emphasizing semantic similarities and differences.

\textit{Cross-Attention Mechanism.} In order to fuse the features in the TS and CS modules at a fine-grained level, we leverage a multi-scale encoder to combine information between two temporal and spatial patterns and produce stronger fusion features for factor generations, motivated by CrossViT \cite{chen2021crossvit}. The architecture incorporates cross-attention layers that explicitly learn correlations between two patterns, enabling the encoder to emphasize mutually informative regions while filtering out noise. This design balances accuracy and computational efficiency, ensuring the fused features yield richer and more discriminative signals for factor generation and that the resulting factors are more representative of market dynamics.

\textit{Contrastive Learning.} To further enhance semantic consistency between TS and CS features, we integrate contrastive learning, drawing inspiration from its success in aligning multimodal representations \cite{li2023blip,radford2021learning}. A contrastive loss is introduced within the factor module to encourage features from TS and CS modules that represent similar market conditions to cluster closely in the embedding space, while pushing apart features corresponding to different conditions. This dual-stream alignment improves the coherence of the factor space and enhances the model’s ability to generalize across diverse market scenarios.

\subsubsection{Prior-Posterior Learning}

Through multi-scale encoder and contrastive learning layer, the CS and TS latent features are fused and aligned. In order to retain the codebook's categorical properties for factors, we incorporate the codebook embeddings $\mathbf{z}_q^{ts}(\mathbf{x})$ and $\mathbf{z}_q^{cs}(\mathbf{x})$ as extra [CLS]-tokens \cite{dosovitskiy2020image}, which are appended to the latent features $\mathbf{z}_e^{ts}(\mathbf{x})$ and $\mathbf{z}_e^{cs}(\mathbf{x})$, respectively. Then, two latent features concat together to get the latent factor embeddings  $\mathbf{z}_e(\mathbf{x})$.

Given the inherent volatility and complexity of the stock market, bridging the gap between noisy market data and an effective factor model remains a significant challenge \cite{duan2022factorvae}. Therefore, we introduce the prior-posterior structure to predict future returns and optimize latent factor embeddings. In the training stage, the posterior layer estimates posterior distribution of factor expected returns $\mathbf{z}_{post}$ from the true future stock returns $\mathbf{y}$ and the latent factor embeddings $\mathbf{z}_e(\mathbf{x})$:
\begin{equation}
    [\mu_{post},\sigma_{post}] = \phi_{FE}(\mathbf{y},\mathbf{z}_e(\mathbf{x})) \text{,}
\end{equation}and $\mathbf{z}_{post}$ follows the independent Gaussian distribution. In the inference stage, the prior layer only use latent factor embeddings without any future information leakage to calculate prior distribution of factor expected returns $\mathbf{z}_{prior}$:
\begin{equation}
    [\mu_{prior},\sigma_{prior}] = \phi_{FP}(\mathbf{z}_e(\mathbf{x})) \text{.}
\end{equation} The return predictor computes future returns based on the factor expected return $\mathbf{z}$:
\begin{equation}
    \hat{\mathbf{y}}=\boldsymbol{\alpha} + \sum_{k=1}^K\boldsymbol{\beta}^k\mathbf{z}^k + \boldsymbol{\epsilon} \text{.}
\end{equation}
The predicted returns $\hat{\mathbf{y}}$ can be directly integrated into portfolio optimization, algorithmic trading strategies, enabling dynamic rebalancing and improved risk management.

\subsection{Downstream Tasks}
\label{sec: Downstream Tasks}
Existing research on latent factor models \cite{duan2022factorvae,wei2023hirevae} primarily focuses on predicting future returns by analyzing cross-sectional characteristics of market stocks, which means the profitability of the factor model in financial investment is typically evaluated through PM tasks. In contrast, \texttt{STORM} captures the cross-sectional features among stocks and the time-series features of individual stocks, enabling it to excel in both PM and single-asset AT tasks.

\subsubsection{Portfolio Management}
To evaluate the performance of the \texttt{STORM} and compare it with baseline methods, we follow the paradigm proposed by FactorVAE \cite{duan2022factorvae}, focusing on return prediction and the PM task. Specifically, we utilize the factor decoder network to generate stock future returns $\mathbf{\hat{y}}$, and then apply the \textit{TopK-Drop} strategy\footnote{https://qlib.readthedocs.io/en/latest/component/strategy.html}, which constructs a daily portfolio by selecting the top $k$ stocks based on predicted returns, to backtest the factor model. Considering transaction costs in real markets, there is a turnover constraint that limits the number of changes in the portfolio composition between consecutive trading days. Specifically, when moving from $\mathbf{w}_t$ to $\mathbf{w}_{t+1}$, at most $d$ stocks can be replaced, where a replacement is defined as a stock whose weight changes from $0$ (not held) to a positive value (newly bought) or from a positive value to $0$ (fully sold). This constraint ensures that at least $k-d$ stocks remain in the portfolio, reducing excessive turnover.

\subsubsection{Algorithmic trading} In \texttt{STORM}, $\mathbf{z}_e(\mathbf{x})$ is the stock latent factor embeddings encoded by the dual VQ-VAE architecture, including both time-series and cross-sectional factor embeddings in the data. The latent factor embeddings $\mathbf{Z}$ are integrated into the observation set $\mathcal{O}=\{\mathbf{Z}, \mathcal{R}\}$, where $\mathcal{R}$ is the reward function used to guide the agent's learning and decision-making in the environment. By combining latent factor embeddings with reward, the agent can more accurately understand the market environment and adjust its strategy based on observations. In the AT task, since it involves trading a single stock, the features of other stocks are excluded. We use the Proximal Policy Optimization (PPO) algorithm \cite{schulman2017proximal} to optimize the policy. 
In each iteration, the agent interacts with the environment to collect trajectories, computes advantage estimates, and updates the policy parameters via PPO's clipped surrogate objective. This optimization process ensures the stability and efficiency of policy updates, helping the agent gradually improve the decision-making ability and achieve better trading strategies.
\begin{table*}[h]
\caption{Comparison of baselines across different tasks}
    \centering
    \renewcommand{\arraystretch}{1.0}
    \resizebox{0.9\linewidth}{!}{
    \begin{threeparttable}
    \begin{tabular}{>{\centering\arraybackslash}p{2.9cm}>
    {\centering\arraybackslash}p{0.01cm} >
    {\centering\arraybackslash}p{0.8cm} >
    {\centering\arraybackslash}p{0.01cm} >
    {\centering\arraybackslash}p{0.6cm} >{\centering\arraybackslash}p{0.6cm} >{\centering\arraybackslash}p{1.2cm} >
    {\centering\arraybackslash}p{0.01cm} >{\centering\arraybackslash}p{0.6cm} >{\centering\arraybackslash}p{0.6cm} >{\centering\arraybackslash}p{0.6cm} >
    {\centering\arraybackslash}p{0.01cm} >{\centering\arraybackslash}p{1.2cm} >{\centering\arraybackslash}p{1.2cm} >{\centering\arraybackslash}p{1.2cm} >
    {\centering\arraybackslash}p{0.01cm} >
    {\centering\arraybackslash}p{1.0cm}
    }
        \toprule
        \multirow{2}{*}{Task} && \multicolumn{1}{c}{Market} && \multicolumn{3}{c}{ML \& DL-based} && \multicolumn{3}{c}{RL-based} && \multicolumn{3}{c}{Factor model} && \multicolumn{1}{c}{Ours} \\
        \cmidrule{3-3}\cmidrule{5-7}\cmidrule{9-11}\cmidrule{13-15}\cmidrule{17-17}
        
        && B\&H && LGBM \cite{yang2020qlib} & LSTM \cite{yu2018forecasting} & Transformer \cite{yang2020qlib} && SAC \cite{haarnoja2018soft} & PPO \cite{schulman2017proximal} & DQN \cite{mnih2013playing} && CAFactor \cite{gu2021autoencoder} & FactorVAE \cite{duan2022factorvae} & HireVAE \cite{wei2023hirevae} && \textbf{\texttt{STORM}} \\
        \midrule
        Prediction  && \cmark && \cmark & \cmark & \cmark && \xmark &  \xmark & \xmark && \cmark & \cmark & \cmark && \cmark \\
        Portfolio Management && \cmark && \cmark & \cmark & \cmark && \cmark & \cmark & \cmark && \cmark & \cmark & \cmark && \cmark \\
        Algorithmic Trading && \cmark && \cmark & \cmark & \cmark && \cmark & \cmark & \cmark && \xmark &  \xmark & \xmark && \cmark  \\
        \bottomrule
    \end{tabular}
    \begin{tablenotes}
    \scriptsize
        \item * {\cmark} indicates that the model is applicable, while {\xmark} signifies that it is \textbf{NOT} applicable to the corresponding task.
    \end{tablenotes}
    \end{threeparttable}
    }
    \label{tab:baseline_comparison}
\end{table*}
\section{Experiment}

In this section, we evaluate the proposed \texttt{STORM} on real stock markets and conduct extensive experiments to address the following research questions. 

\begin{description}[leftmargin=1.5em,labelindent=0em,align=left]
    \item[\textbf{RQ1}] How does \texttt{STORM} perform on downstream tasks?
    \item[\textbf{RQ2}] How to evaluate the effectiveness of \texttt{STORM}'s learned factors?
    \item[\textbf{RQ3}] How do the key components contribute to the performance of \texttt{STORM}?
\end{description}

\subsection{Experiment Settings}

\subsubsection{Datasets}

We conduct experiments on two U.S. stock markets, SP500 and DJ30, using stock daily data that includes technical features based on \textit{Alpha158} \cite{yang2020qlib}. Both datasets span 16 years, from 2008-04-01 to 2024-03-31, encompassing global conditions, e.g., the 2007-2008 financial crisis and COVID-19. Datasets are chronologically divided into non-overlapping training (from 2008-04-01 to 2021-03-31) and test (from 2021-04-01 to 2024-03-31) sets.

\subsubsection{Metrics}

We compare \texttt{STORM} and baselines in terms of 6 financial metrics across the PM and AT tasks. The financial metrics include 2 profit criteria, annualized percentage yield (APY) and Cumulative Wealth (CW), 2 risk-adjusted profit criteria, Calmar ratio (CR) and annualized Sharpe ratio (ASR), and 2 risk criteria, maximum drawdown (MDD) and annualized volatility (AVO). The calculation formulas and meanings of these metrics are as follows:
\begin{itemize}[left=0.3em]
    \item \textbf{CW} is the total returns yielded from a portfolio strategy: $\text{CW}_T = \prod_{i=1}^{T}(1+r_i)$, where $r_i$ is the net return. 

    \item \textbf{APY}  measures the average
wealth increment that one portfolio strategy could achieve compounded in a year, which is defined as $\text{APY}_T = \sqrt[y]{\text{CW}_T} - 1$, where $y$ is the number of years corresponding to $T$ trading rounds.

    \item \textbf{MDD} measures the largest loss from a historical peak in the cumulative wealth to show the worst case, which is defined as: $MDD = \mathop{\max}_{i=0}^T \frac{P_i-R_i}{P_i}$, where $R_i = \prod_{i=1}^{T} {\frac{V_i}{V_{i-1}}} $ and $ P_i = \mathop{\max}_{i=1}^T R_i$.

    \item \textbf{AVO} is the annualized standard deviation of daily returns and multiplied by $\sqrt{AT}$, where $AT$ is the average trading rounds of annual trading days and $AT=252$ for all the datasets.

    \item \textbf{ASR} is an annualized volatility risk-adjusted return defined as $\text{ASR} = \frac{\text{APY}-R_f}{\text{AVO}}$, where $R_f$ is the risk-free return.

    \item  \textbf{CR} measures the drawdown risk-adjusted return of a portfolio calculated as $CR=\frac{\text{APY}}{\text{MDD}}$.
\end{itemize}

Typically, to evaluate the effectiveness of the learned factors in the PM task, we adopt the Rank Information Coefficient (RankIC) and the Information Ratio of RankIC (RankICIR):
\begin{itemize}[left=0.3em]
    \item \textbf{RankIC} is a ranking metric in finance, which measures the correlation between the predicted rankings and the actual returns. It is defined as:
\begin{equation*}
    \begin{aligned}
        & \text{RankIC}_s = \frac{1}{N} \frac{(r_{\hat{y}_s} -\text{mean}(r_{\hat{y}_s}))^T(r_{y_t} - \text{mean}(r_{y_s}))}{\text{std}(r_{\hat{y}_s}) \cdot \text{std}(r_{y_s})} \\ 
        & \text{RankIC} = \frac{1}{T_{\text{test}}} \sum_{s=1}^{T_{\text{test}}}\text{RankIC}_s \text{,}\\
    \end{aligned}
\end{equation*}where $T_{test}$ denotes the number of trading days in the test range, $r_{y_s}$ and $r_{\hat{y}_s}$ represent the true and predicted ranks of stocks on the trading day $s$, respectively.

    \item \textbf{RankICIR} is the information ratio of RankIC, which measures the stability of prediction,
    \begin{equation*}
        \text{RankICIR} = \frac{\text{mean}(\text{RankIC}_s)}{\text{std}(\text{RankIC}_s)} \text{.}
    \end{equation*}
\end{itemize} 

\begin{table}[b]
\caption{Portfolio management task results of all models across six metrics (mean ± range, computed across 10 runs).}
\renewcommand{\arraystretch}{0.5}
\setlength{\abovecaptionskip}{0.3cm}
\centering
\begin{threeparttable}
\small
\setlength{\tabcolsep}{0.8mm}{
\resizebox{0.9\linewidth}{!}{
\begin{tabular}{lccccccccc} 
\toprule
\multicolumn{10}{c}{\textbf{SP500 Dataset}}\\
\midrule

\multirow{3}{*}{\begin{tabular}[c]{@{}c@{}}Strategies\end{tabular}} && \multicolumn{2}{c}{Profit} && \multicolumn{2}{c}{Risk-Adj. Profit} && \multicolumn{2}{c}{Risk} \\ 

\cmidrule{3-4}\cmidrule{6-7}\cmidrule{9-10}

&& APY$\uparrow$ & CW$\uparrow$ && CR$\uparrow$ & ASR$\uparrow$ && MDD$\downarrow$ & AVO$\downarrow$ \\
\midrule

Market Index && 0.058 & 1.184 && 0.228 & 0.142 && 0.254 & 0.410 \\
\midrule

\multirow{2}{*}{LightGBM} 	&&	0.059	&	1.201	&&	0.304 &	0.332	&&	0.238	&	0.176	\\
&& \scriptsize ± 0.132	&	\scriptsize ± 0.456	&&	\scriptsize ± 0.753	&	\scriptsize ± 0.755	&&	\scriptsize ± 0.108	&	\scriptsize ± 0.008	\\

\multirow{2}{*}{LSTM} 	&& 0.069 &	1.221	&&	0.278 &	0.371	&&	0.248	&	0.186	\\
&& \scriptsize ± 0.020	&	\scriptsize ± 0.068	&&	\scriptsize ± 0.065	&	\scriptsize ± 0.114	&&	\scriptsize ± 0.042	&	\scriptsize ± 0.013	\\

\multirow{2}{*}{Transformer}	&& 0.076&	1.244	&&	0.389 &	0.433	&&	\underline{0.198}\footnotemark[1]	&	0.174\\

&& \scriptsize ± 0.028	&	\scriptsize ± 0.098	&&	\scriptsize ± 0.165	&	\scriptsize ± 0.161	&&	\scriptsize ± 0.057	&	\scriptsize ± 0.004	\\

\multirow{2}{*}{CAFactor}	&& 0.075	&	1.241	&&	0.342	&	0.428	&&	0.223	&	0.174	\\
&& \scriptsize ± 0.028	&	\scriptsize ± 0.100	&&	\scriptsize ± 0.229	&	\scriptsize ± 0.165	&&	\scriptsize ± 0.043	&	\scriptsize ± 0.006	\\

\multirow{2}{*}{FactorVAE}	&& \underline{0.079}	&	\underline{1.256}	&&	\underline{0.404}	&	\underline{0.460}	&&	0.200	&	0.173	\\
&& \scriptsize ± 0.025	&	\scriptsize ± 0.085	&&	\scriptsize ± 0.177	&	\scriptsize ± 0.145	&&	\scriptsize ± 0.040	&	\scriptsize ± 0.007	\\

\multirow{2}{*}{HireVAE}	&& 0.077 &	1.249	&&	0.361	&	0.448	&&	0.216	&	\underline{0.172}	\\
&& \scriptsize ± 0.029	&	\scriptsize ± 0.104	&& \scriptsize ± 0.180	&	\scriptsize ± 0.189	&&	\scriptsize ± 0.048	&	\scriptsize ± 0.007	\\
\midrule
\multirow{2}{*}{\textbf{\texttt{STORM}}}	&& \textbf{0.188}\footnotemark[2] &	\textbf{1.683}	&&	\textbf{1.189}	&	\textbf{1.052}	&&	0.166	&	0.171	\\
&& \scriptsize ± 0.055 &	\scriptsize ± 0.226	&&	\scriptsize ± 0.661 	&	\scriptsize ± 0.329	&&	\scriptsize ± 0.050	&	\scriptsize ± 0.020	\\

\multirow{2}{*}{\texttt{STORM}-w/o-TS}	&& 0.090  &	1.300  &&  0.503  &	0.570	&&	0.181 	& 0.175	\\
&& \scriptsize ± 0.074	&	\scriptsize ± 0.277	&& \scriptsize ± 0.390	&	\scriptsize ± 0.339 	&&	\scriptsize ± 0.043  & \scriptsize ± 0.025	\\

\multirow{2}{*}{\texttt{STORM}-w/o-CS}	&&  0.089	&	1.294	&&	0.623	&	0.592  && \textbf{0.146} & \textbf{0.167}  \\
&& \scriptsize ± 0.045	&	\scriptsize ± 0.163	&&	 \scriptsize ± 0.385 	&	\scriptsize ± 0.252	&& \scriptsize ± 0.073	&	\scriptsize ± 0.020	\\

\midrule
Improvement(\%)\footnotemark[3] &&  137.97	&	34.00	&&	194.31	&	128.70 	&&	26.26	& 2.91 	\\
\midrule
\midrule
\multicolumn{10}{c}{\textbf{DJ30 Dataset}}\\
\midrule

\multirow{3}{*}{\begin{tabular}[c]{@{}c@{}}Strategies\end{tabular}} && \multicolumn{2}{c}{Profit} && \multicolumn{2}{c}{Risk-Adj. Profit} && \multicolumn{2}{c}{Risk} \\ 

\cmidrule{3-4}\cmidrule{6-7}\cmidrule{9-10}
&& APY$\uparrow$ & CW$\uparrow$ && CR$\uparrow$ & ASR$\uparrow$ && MDD$\downarrow$ & AVO$\downarrow$ \\
\midrule
Market Index && 0.063  & 1.201 && 0.147 & 0.429 && 0.219 & 0.288\\
\midrule

\multirow{2}{*}{LightGBM} && 0.069 & 1.221 &&	0.288 &	0.430 &&	0.244	&	0.160  \\
&& \scriptsize ± 0.040	&	\scriptsize ± 0.140	&& \scriptsize ± 0.262	&	\scriptsize ± 0.267	&&	\scriptsize ± 0.047	&	\scriptsize ± 0.007 \\

\multirow{2}{*}{LSTM} &&	0.060	&	1.192	&&	0.243 &	0.370	&&	0.248 &	0.163 \\
&&	\scriptsize ± 0.007	&	\scriptsize ± 0.023	&&	\scriptsize ± 0.021	& \scriptsize ± 0.043	&&	\scriptsize ± 0.006	&	\scriptsize ± 0.005 \\

\multirow{2}{*}{Transformer} && 0.056	&	1.179	&&	0.227	&	0.367	&&	0.250&	0.154 \\
&& \scriptsize ± 0.013	&	\scriptsize ± 0.043	&&	\scriptsize ± 0.067	&	\scriptsize ± 0.085	&&	\scriptsize ± 0.033	& \scriptsize ± 0.002 \\

\multirow{2}{*}{CAFactor} 	&& 0.059	&	1.186	&&	0.233	&	0.382 &&	0.252	&	\underline{0.153}  \\
&& \scriptsize ± 0.015	&	\scriptsize ± 0.050	&& \scriptsize ± 0.058	&	\scriptsize ± 0.092	&&	\scriptsize ± 0.021	&	\scriptsize ± 0.001  \\

\multirow{2}{*}{FactorVAE}	&& \underline{0.076}	&	\underline{1.246}	&&	\underline{0.352} & \underline{0.480}	&&	\underline{0.225}	& 0.159 \\
&& \scriptsize ± 0.038	&	\scriptsize ± 0.135	&&	\scriptsize ± 0.349	&	\scriptsize ± 0.235	&&	\scriptsize ± 0.063 &	\scriptsize ± 0.005 \\

\multirow{2}{*}{HireVAE} &&	0.072	&	1.233	&&	0.298&	0.445	&&	0.247&	0.163\\

&&	\scriptsize ± 0.037	& \scriptsize ± 0.132	&&	\scriptsize ± 0.253	&	\scriptsize ± 0.256 &&	\scriptsize ± 0.057	&	\scriptsize ± 0.012 \\
\midrule

\multirow{2}{*}{\textbf{\texttt{STORM}}} && \textbf{0.148} 	&	\textbf{1.517}	&&	\textbf{1.396}	& \textbf{1.052}  &&	\textbf{0.108}  & \textbf{0.140}  \\
&& \scriptsize ± 0.046 	&	\scriptsize ± 0.188	&&	 \scriptsize ± 0.679	&	\scriptsize ± 0.297	&& \scriptsize ± 0.026	&	\scriptsize ± 0.014	\\

\multirow{2}{*}{\texttt{STORM}-w/o-TS}	&&  0.079	&	1.259 	&&	0.621	& 0.603 &&	0.127 &	0.142 \\
&& \scriptsize ± 0.057	&	\scriptsize ± 0.192	&&	 \scriptsize ± 0.425	&	\scriptsize ± 0.379	&& \scriptsize ± 0.038	&	\scriptsize ± 0.008	\\

\multirow{2}{*}{\texttt{STORM}-w/o-CS}	&&  0.073	&	1.236	&&	0.533	& 0.566 &&	0.138 &	\textbf{0.140}  \\
&& \scriptsize ± 0.051	&	\scriptsize ± 0.169	&&	 \scriptsize ± 0.400	&	\scriptsize ± 0.343	&& \scriptsize ± 0.034	&	\scriptsize ± 0.017	\\
\midrule
Improvement(\%) &&  94.74	&	21.75 	&&	296.59	& 119.17 	&&	52.00	&	8.50	\\
\bottomrule

\end{tabular}
}
\begin{tablenotes}
\scriptsize
\item[$1$] \underline{Underline} indicates the best-performing baseline method result;
\item[$2$] \textbf{Bold} indicates the best performance among our proposed models (including variants);
\item[$3$] Improvement of \texttt{STORM} over the best-performing baselines.
\end{tablenotes}
}
\end{threeparttable}
\label{table:app_baseline_result}
\end{table}

\begin{table*}[ht]
\caption{Algorithmic trading task results on all models across six standard metrics (averaged over 5 runs).}
\renewcommand{\arraystretch}{0.8}
\setlength{\abovecaptionskip}{0.1cm}
\footnotesize
\centering
\resizebox{0.9\textwidth}{!}{ 
\begin{tabular}{lccclccclccclccclccc}
\toprule
\multirow{3}{*}{Models} & \multicolumn{3}{c}{AAPL} & & \multicolumn{3}{c}{JPM} & & \multicolumn{3}{c}{IBM} & & \multicolumn{3}{c}{INTC} & & \multicolumn{3}{c}{MSFT}   \\ 
\cmidrule{2-4}\cmidrule{6-8}\cmidrule{10-12}\cmidrule{14-16}\cmidrule{18-20}

& APY$\uparrow$  & CW$\uparrow$  & CR$\uparrow$  & & APY$\uparrow$  & CW$\uparrow$  & CR$\uparrow$ & & APY$\uparrow$  & CW$\uparrow$  & CR$\uparrow$ & & APY$\uparrow$  & CW$\uparrow$  & CR$\uparrow$ & & APY$\uparrow$  & CW$\uparrow$  & CR$\uparrow$  \\ 
\midrule
Buy\&Hold & 0.120 	&	1.404 	&	0.383 	&&	0.096 	&	1.316 	&	0.236 	&&	0.145 	&	1.499 	&	0.727 	&&	-0.117	&	0.690 	&	-0.184	&&	0.214 	&	1.784 	&	0.569 	\\
\midrule

LightGBM & 0.135	&	1.390 	&	0.487 	&&	0.116 	&	1.335 	&	0.333 	&&	\underline{0.227} 	&	\underline{1.654} 	&	1.091 	&&	-0.042	&	0.880 	&	0.038 	&&	\underline{0.267} 	&	\underline{2.068} 	&	0.637 	\\

\multirow{2}{*}{LSTM} & 0.053 	&	1.152 	&	0.283 	&&	0.079 	&	1.290  	&	0.266  &&	0.134 	&	1.386 	&	0.754 	&&	0.060 	&	1.262 	&	0.381 	&&	0.178 	&	1.513 	&	0.893 	\\
& \scriptsize ± 0.015	&	\scriptsize ± 0.042	&	\scriptsize ± 0.101	&&	\scriptsize ± 0.019	&	\scriptsize ± 0.053	&	\scriptsize± 0.023	&&	\scriptsize ± 0.141	&	\scriptsize ± 0.406	&	\scriptsize ± 0.69	&&	\scriptsize ± 0.119	&	\scriptsize ± 0.429	&	\scriptsize ± 0.62	&&	\scriptsize ± 0.122	&	\scriptsize ± 0.352	&	\scriptsize ± 0.822	\\

\multirow{2}{*}{Transformer} & 0.083 	&	1.240 	&	0.512 	&&	\underline{0.133} 	&	\underline{1.384}  	&	\underline{0.614} &&	0.131 	&	1.377 	&	0.782 	&&	\underline{0.079} 	&	\underline{1.290} 	&	\underline{0.458} &&	0.138 	&	1.397 	&	0.726 	\\
& \scriptsize ± 0.099	&	\scriptsize ± 0.284	&	\scriptsize ± 0.864	&&	\scriptsize ± 0.07	&	\scriptsize ± 0.201	&	\scriptsize ± 0.509	&&	\scriptsize ± 0.072	&	\scriptsize ± 0.207	&	\scriptsize ± 0.541	&&	\scriptsize ± 0.174	&	\scriptsize ± 0.68	&	\scriptsize ± 0.824	&&	\scriptsize± 0.118	&	\scriptsize ± 0.338	&	\scriptsize ± 0.643	\\

\multirow{2}{*}{DQN} &	0.135 	&	1.374 	&	0.510 	&&	0.105 	&	1.305 	&	0.607  	&&	0.139 	&	1.400 	&	0.802 	&&	0.061 	&	1.185 	&	0.442 	&&	0.166 	&	1.475 	&	0.534 	\\
& \scriptsize ± 0.075	&	\scriptsize ± 0.226	&	\scriptsize ± 0.239	&&	\scriptsize ± 0.132	&	\scriptsize ± 0.383	&	\scriptsize ± 0.673	&&	\scriptsize ± 0.055	&	\scriptsize ± 0.158	&	\scriptsize ± 0.270	&&	\scriptsize ± 0.106	&	\scriptsize ± 0.305	&	\scriptsize ± 0.436	&&	\scriptsize ± 0.035	&	\scriptsize ± 0.100	&	\scriptsize ± 0.107	\\

\multirow{2}{*}{SAC} & \underline{0.147}	&	\underline{1.509} 	&	\underline{0.528} 	&&	0.131 	&	1.383 	&	0.400 	&&	0.207 	&	1.598 	&	\underline{1.170} 	&&	0.056 	&	1.165 	&	0.353 	&&	0.229 	&	1.656 	&	\underline{0.929} 	\\
& \scriptsize ± 0.021	&	\scriptsize ± 0.060	&	\scriptsize ± 0.052	&&	\scriptsize ± 0.038	&	\scriptsize ± 0.111	&	\scriptsize ± 0.198	&&	\scriptsize ± 0.057	&	\scriptsize ± 0.163	&	\scriptsize ± 0.335	&&	\scriptsize ± 0.210	&	\scriptsize ± 0.6	&	\scriptsize ± 0.994	&&	\scriptsize ± 0.114	&	\scriptsize ± 0.331	&	\scriptsize ± 0.585	\\

\multirow{2}{*}{PPO} & 0.137 	&	1.379 	&	0.496 	&&	0.128 	&	1.372 	&	0.356 	&&	0.146 	&	1.422 	&	0.779 	&&	-0.019	&	0.954 	&	0.040 	&&	0.216 	&	1.620 	&	0.569 	\\
& \scriptsize  ± 0.073	&	\scriptsize  ± 0.209	&	\scriptsize  ± 0.213	&&	\scriptsize  ± 0.026	&	\scriptsize  ± 0.071	&	\scriptsize  ± 0.059	&&	\scriptsize  ± 0.047	&	\scriptsize  ± 0.136	&	\scriptsize  ± 0.205	&&	\scriptsize  ± 0.177	&	\scriptsize  ± 0.514	&	\scriptsize  ± 0.418	&&	\scriptsize  ± 0.081	&	\scriptsize  ± 0.232	&	\scriptsize ± 0.105	\\

\midrule

\multirow{2}{*}{\textbf{\texttt{STORM}}} &  \textbf{0.229} 	& \textbf{1.857}  & 0.750  && \textbf{0.174} 	&  \textbf{1.621}	& 0.559  && \textbf{0.236}	& \textbf{1.893}	& \textbf{1.470} 	&& \textbf{0.173}	& \textbf{1.625} 	&  \textbf{0.773}	&& \textbf{0.290}  & \textbf{2.154} 	& \textbf{1.216}  
  	\\
& \scriptsize ± 	0.033 	& \scriptsize ± 	0.154 	& \scriptsize ± 	0.066 	&&	\scriptsize ± 	0.032 	& \scriptsize ± 	0.133 	& \scriptsize ± 	0.081 	&&	\scriptsize ± 	0.039 	& \scriptsize ± 	0.184 	& \scriptsize ± 	0.445 	&&	\scriptsize ± 	0.067 	& \scriptsize ± 	0.284  	& \scriptsize ± 	0.293  	&&	\scriptsize ± 	0.052 	& \scriptsize ± 	0.262 	& \scriptsize ± 	0.597 
	\\

\multirow{2}{*}{\texttt{STORM}-w/o-TS}& 0.199 	&	1.730 	&	\textbf{0.766}	&&	0.127 	&	1.437 	&	\textbf{0.627}  	&&	0.154 	&	1.536 	&	1.294 	&&	0.106	&	1.212 	&	0.267  	&&	0.254 	&	1.979 	&	0.964 	\\
& \scriptsize ± 0.059	&	\scriptsize ± 0.260	&	\scriptsize ± 0.274	&&	\scriptsize ± 0.058	&	\scriptsize ± 0.214	&	\scriptsize ± 0.187	&&	\scriptsize± 0.023	&	\scriptsize ± 0.089	&	\scriptsize ± 1.012	&&	\scriptsize ± 0.120	&	\scriptsize ± 0.145	&	\scriptsize ± 0.135	&&	\scriptsize ± 0.060	&	\scriptsize ± 0.093	&	\scriptsize ± 0.503	\\

\multirow{2}{*}{\texttt{STORM}-w/o-CS} & 0.148 	&	1.521 	&	0.641 	&&	0.103 	&	1.347 	&	0.350 	&&	0.152 	&	1.534 	&	0.814 	&&	0.136	&	1.474 	&	0.541 	&&	0.181 	&	1.650 	&	0.648 	\\
& \scriptsize ± 0.069	&	\scriptsize ± 0.283	&	\scriptsize ± 0.184	&&	\scriptsize ± 0.043	&	\scriptsize ± 0.154	&	\scriptsize ± 0.106	&&	\scriptsize± 0.054	&	\scriptsize ± 0.219	&	\scriptsize ± 0.352	&&	\scriptsize± 0.082	&	\scriptsize ± 0.300	&	\scriptsize ± 0.328	&&	\scriptsize ± 0.051	&	\scriptsize ± 0.221	&	\scriptsize ± 0.361	\\
\midrule

Improvement(\%) &	55.782 	&	23.062 	&	45.076	&&	30.827 	&	17.124 	&	2.117 	&&	3.965 	&	14.4501 	&	20.408 	&&	118.987 	&	26.969 	&	66.594  	&&	8.614 	&	4.159 	&	30.893 
 \\
\bottomrule
\toprule
 \multirow{3}{*}{Models} & \multicolumn{3}{c}{AAPL} && \multicolumn{3}{c}{JPM} && \multicolumn{3}{c}{IBM} & & \multicolumn{3}{c}{INTC} & & \multicolumn{3}{c}{MSFT}   \\ 
\cmidrule{2-4}\cmidrule{6-8}\cmidrule{10-12}\cmidrule{14-16}\cmidrule{18-20}
& ASR$\uparrow$  & MDD$\downarrow$ & AVO$\downarrow$  && ASR$\uparrow$  & MDD$\downarrow$ & AVO$\downarrow$  && ASR$\uparrow$  & MDD$\downarrow$ & AVO$\downarrow$ && ASR$\uparrow$  & MDD$\downarrow$ & AVO$\downarrow$  && ASR$\uparrow$  & MDD$\downarrow$ & AVO$\downarrow$  \\ 
\midrule

Buy\&Hold & 0.447 	&	0.313 	&	0.269 	&&	0.405 	&	0.406 	&	0.237 	&&	0.679 	&	0.199 	&	0.213 	&&	-0.324	&	0.635 	&	0.359 	&&	0.777 	&	0.376 	&	0.275 	\\
\midrule

LightGBM & 0.950 	&	0.309 	&	0.017 	&&	0.921 	&	0.386 	&	0.015 	&&	\underline{1.822} 	&	0.175 	&	0.011 	&&	0.098 	&	0.553 	&	0.023 	&&	1.456 	&	0.370 	&	0.017 	\\

\multirow{2}{*}{LSTM} & 0.623 	&	0.244 	&	0.012 	&&	0.801 	&	0.335 	&	0.012 	&&	1.268 	&	0.163 	&	0.010 	&&	0.287 	&	\underline{0.149} 	&	\underline{0.010}	&&	1.345 	&	0.246 	&	0.014 	\\
& \scriptsize ± 0.163	&	\scriptsize ± 0.072	&	\scriptsize ± 0.003	&&	\scriptsize ± 0.185	&	\scriptsize ± 0.061	&	\scriptsize ± 0.002	&&	\scriptsize ± 0.924	&	\scriptsize ± 0.017	&	\scriptsize ± 0.001	&&	\scriptsize ± 1.02	&	\scriptsize ± 0.101	&	\scriptsize± 0.007	&&	\scriptsize ± 0.842	&	\scriptsize ± 0.124	&	\scriptsize ± 0.003	\\

\multirow{2}{*}{Transformer} & 0.972 	&	\underline{0.239} 	&	\underline{0.010} 	&&	\underline{1.332 }	&	0.226 	&	\underline{0.010} 	&&	1.336 	&	0.182 	&	0.010 	&&	0.604 	&	0.257 	&	0.014 	&&	1.206 	&	\underline{0.182} 	&	\underline{0.011} 	\\
& \scriptsize ± 0.748	&	\scriptsize ± 0.187	&	\scriptsize ± 0.006	&&	\scriptsize ± 0.607	&	\scriptsize ± 0.100	&	\scriptsize ± 0.003	&&	\scriptsize ± 0.688	&	\scriptsize ± 0.100	&	\scriptsize± 0.003	&&	\scriptsize ± 0.789	&	\scriptsize ± 0.194	&	\scriptsize ± 0.006	&&	\scriptsize ± 0.754	&	\scriptsize ± 0.134	&	\scriptsize ± 0.006	\\

\multirow{2}{*}{DQN} & 0.974 	&	0.288 	&	0.016 	&&	1.085 	&	\underline{0.212} 	&	\underline{0.010 } &&	1.415 	&	0.161 	&	\underline{0.009} 	&&	\underline{0.609} 	&	0.243 	&	0.014 	&&	1.194 	&	0.309 	&	0.015 	\\
& \scriptsize ± 0.472	&	\scriptsize ± 0.159	&	\scriptsize ± 0.007	&&	\scriptsize ± 1.270	&	\scriptsize ± 0.118	&	\scriptsize ± 0.003	&&	\scriptsize ± 0.403	&	\scriptsize ± 0.049	&	\scriptsize ± 0.004	&&	\scriptsize ± 0.584	&	\scriptsize ± 0.310	&	\scriptsize ± 0.009	&&	\scriptsize ± 0.148	&	\scriptsize ± 0.079	&	\scriptsize ± 0.004	\\

\multirow{2}{*}{SAC} & \underline{1.054} 	&	0.292 	&	0.016 	&&	1.044 	&	0.350 	&	0.014 	&&	1.794 	&	\underline{0.157} 	&	0.011 	&&	0.527 	&	0.359 	&	0.017 	&&	\underline{1.638} 	&	0.229 	&	0.013 	\\
& \scriptsize ± 0.223	&	\scriptsize ± 0.022	&	\scriptsize ± 0.002	&&	\scriptsize ± 0.276	&	\scriptsize ± 0.085	&	\scriptsize ± 0.001	&&	\scriptsize ± 0.458	&	\scriptsize ± 0.068	&	\scriptsize ± 0.005	&&	\scriptsize ± 1.061	&	\scriptsize ± 0.191	&	\scriptsize ± 0.004	&&	\scriptsize ± 0.736	&	\scriptsize ± 0.215	&	\scriptsize ± 0.005	\\

\multirow{2}{*}{PPO} & 0.974 	&	0.300 	&	0.016 	&&	1.000 	&	0.381 	&	0.015 	&&	1.386 	&	0.177 	&	0.011 	&&	-0.044	&	0.511 	&	0.019 	&&	1.276 	&	0.353 	&	0.017 	\\
& \scriptsize  ± 0.412	&	\scriptsize  ± 0.030	&	\scriptsize  ± 0.001	&&	\scriptsize  ± 0.189	&	\scriptsize  ± 0.018	&	\scriptsize  ± 0.001	&&	\scriptsize  ± 0.231	&	\scriptsize  ± 0.027	&	\scriptsize  ± 0.003	&&	\scriptsize  ± 1.689	&	\scriptsize  ± 0.115	&	\scriptsize  ± 0.007	&&	\scriptsize  ± 0.299	&	\scriptsize  ± 0.036	&	\scriptsize  ± 0.001	\\

\midrule
\multirow{2}{*}{\textbf{\texttt{STORM}}}& \textbf{1.346}	& 0.282 & 0.016 	&& 1.245 	& 0.299 & 0.014	&& \textbf{2.107} & 0.137 & 0.009	&& \textbf{1.189} 	& \textbf{0.227} & \textbf{0.015} 	&& \textbf{1.672}	& \textbf{0.210}	& \textbf{0.015} 
 	\\
& \scriptsize ± 0.162 	& \scriptsize ± 0.000 	& \scriptsize ± 0.001 	
&&	\scriptsize ± 0.106 	& \scriptsize ± 0.016 	& \scriptsize ± 0.001 
&&	\scriptsize ± 0.302 	& \scriptsize ± 0.033 	& \scriptsize ± 0.001 
&&	\scriptsize ± 0.386  	& \scriptsize ± 0.055 	& \scriptsize ± 0.002 
&&	\scriptsize ± 0.201 	& \scriptsize ± 0.067 	& \scriptsize ± 0.002	\\

\multirow{2}{*}{\texttt{STORM}-w/o-TS}& 1.199 	&	0.262 	&	0.017 	&&	\textbf{1.587} 	&	\textbf{0.194} 	& \textbf{0.011}  	&&	1.868 	&	\textbf{0.131} 	&	\textbf{0.008} 	&&	0.580 	&	0.344 	&	0.017 	&&	1.553 	&	0.244 	&	\textbf{0.015} 	\\
& \scriptsize ± 0.300	&	\scriptsize ± 0.052	&	\scriptsize± 0.001	&&	\scriptsize ± 0.296	&	\scriptsize ± 0.038	&	\scriptsize ± 0.001	&&	\scriptsize ± 0.956	&	\scriptsize ± 0.077	&	\scriptsize ± 0.003	&&	\scriptsize ± 0.229	&	\scriptsize ± 0.061	&	\scriptsize ± 0.000	&&	\scriptsize ± 0.187	&	\scriptsize ± 0.093	&	\scriptsize ± 0.004	\\

\multirow{2}{*}{\texttt{STORM}-w/o-CS} & 1.027 	&	\textbf{0.229} 	&	\textbf{0.015}	&&	0.858 	&	0.327 	&	0.013 	&&	1.397  &	0.177 	&	0.010 	&&	0.914 	&	0.312 	&	0.018 	&&	1.401 	&	0.295 	&	0.016 	\\
& \scriptsize ± 0.326	&	\scriptsize± 0.030	&	\scriptsize ± 0.001	&&	\scriptsize ± 0.236	&	\scriptsize ± 0.064	&	\scriptsize ± 0.001	&&	\scriptsize ± 0.405	&	\scriptsize ± 0.028	&	\scriptsize ± 0.001	&&	\scriptsize± 0.397	& \scriptsize ± 0.136	&	\scriptsize ± 0.004	&&	\scriptsize ± 0.224	&	\scriptsize ± 0.090	&	\scriptsize ± 0.002	\\
\midrule
Improvement(\%)
&	27.704 	&	4.184 	&	- 	&&	19.144 	&	8.491 	&	- 	&&	15.642 	&	16.561 	&	11.111 	&&	95.238 	&	- 	&	-  	&&	2.076 	&	- 	&	-
	\\
\bottomrule
\end{tabular}
}
\label{tab:trading}
\end{table*}

\subsubsection{Baselines}

As shown in Table \ref{tab:baseline_comparison}, we compare \texttt{STORM} with nine methods across one prediction task and two downstream tasks. The baseline methods fall into three categories: ML \& DL-based models, Factor models, and RL-based models. We evaluate \texttt{STORM} against ML\&DL models and factor models in prediction and PM tasks, and against ML\&DL models and RL methods in the AT task. Moreover, \texttt{STORM} is a versatile all-rounder that supports three tasks, unlike other methods specialized in specific tasks.
\begin{itemize}[left=0.3em, itemsep=0.6em]
    \item \textbf{Market}
        \begin{itemize}
            \item Buy-and-Hold (B\&H) involves holding assets for an extended period, regardless of short-term market fluctuations, assuming that long-term returns will be more favorable.
        \end{itemize}
    \item \textbf{ML\&DL-based}
        \begin{itemize}
            \item  LGBM \cite{yang2020qlib} uses a series of tree models to predict price fluctuations and provide buy and sell signals.
            \item LSTM \cite{yu2018forecasting} utilizes long short-term memory to improve the accuracy of price predictions.
            \item Transformer \cite{yang2020qlib} models leverage self-attention mechanisms to enhance the precision of price forecasts.
        \end{itemize}
        
    \item \textbf{RL-based}
        \begin{itemize}
            \item SAC \cite{haarnoja2018soft} is an off-policy actor-critic algorithm using entropy regularization and soft value functions.
            \item PPO \cite{schulman2017proximal} updates trading policies iteratively to balance exploration and exploitation, ensuring stability and efficiency.
            \item DQN \cite{mnih2013playing} approximates the action-value function and make trading decisions from market data with a deep Q-network.
        \end{itemize}
        
    \item \textbf{Factor-model}
        \begin{itemize}
            \item CAFactor \cite{gu2021autoencoder} learns latent factors from asset features.
            \item FactorVAE \cite{duan2022factorvae} is a state-of-the-art VAE-based latent factor model that learns optimal latent stock factors.
            \item HireVAE \cite{wei2023hirevae} is a state-of-the-art VAE-based latent factor model capturing market and inter-stock factors.
        \end{itemize}
\end{itemize}

\subsubsection{Implementation Details} 

All experiments are implemented using PyTorch and conducted on two NVIDIA RTX H100 GPUs with 80GB of memory each. The input data is organized into batches with dimensions $(B, W, N, D)$, where $B = 16$ is the batch size, $N$ is the number of stocks (28 for DJ30, 408 for SP500), $W = 64$ denotes historical days, and $D = 152$ is the total number of features, including the OHLC data and technical indicators.

The TS module uses patches of size $(4,1,D)$ (4-day sequences per stock), whereas the CS module uses $(1,N,D)$ (all stocks on a given day). Encoders (4 layers, 4 heads) and decoders (2 layers, 8 heads) are standard Transformer blocks with a shared embedding size of 256. Codebook sizes are searched in ${128,256,512,1024,2048}$, with 512 performing best.

We adopt AdamW (lr${=}1\times10^{-4}$, weight decay${=}0.05$) with a linear warmup of 100 epochs and a decay to zero over 1000 epochs. The clip loss and TS/CS reconstruction losses are weighted by $1\times10^{-3}$; commitment loss is 1.0; orthogonality, diversity, price prediction, and KL losses are weighted by 0.1.

For downstream tasks, we use Qlib’s\footnote{https://github.com/microsoft/qlib} TopK-Drop portfolio strategy ($k{=}5$, $d{=}3$) and PPO for trading. The pretrained TS/CS factors serve as states; all CS factors are retained while only TS factors relevant to traded stocks are used. The action space is {buy, hold, sell}. Rewards follow $r=\frac{v_{\text{post}}-v_{\text{pre}}}{v_{\text{pre}}}$ with $v=p\times m + c$, starting from $10^6$ initial cash and transaction cost $10^{-4}$.

The implementations of the ML \& DL methods are based on Qlib \cite{yang2020qlib}. As for other baselines, we adopt the default settings in their public implementations. Each experiment is repeated multiple times with different random seeds, and the mean and variance of the metrics are reported. 

\vfill\eject

\subsection{Performance on Investment Tasks (for RQ1)}

We use three types of metrics: profit, risk, and risk-adjusted profit to evaluate \texttt{STORM}'s investment performance on downstream tasks, i.e., portfolio management and algorithmic trading. The evaluation results are presented in Table~\ref{table:app_baseline_result} and Table~\ref{tab:trading}.

For \textit{profitability}, \texttt{STORM} outperforms all the baselines, with average improvements of 116.36\% on APY over the best-performing baseline in the PM task and 43.64\% over the best baseline methods on five stocks in the AT task. These results underscore the model’s capability to extract meaningful factors, and optimize decision-making for enhanced financial performance.
For \textit{risk resistance}, although \texttt{STORM} does not outperform the best baseline methods in terms of MDD and AVO on the INTC and MSFT datasets for the trading task, this can be attributed to the extreme price fluctuations during the COVID-19 event. Both stocks experienced a sharp decline of over 30\%, followed by a subsequent rebound. Given \texttt{STORM}'s strong profitability capability, its trading strategy tends to aggressively capture trends, which naturally leads to larger drawdowns and increased volatility as a trade-off for higher potential returns.

Moreover, considering the \textit{return and risk} simultaneously, \texttt{STORM} performs the best on ASR and CR with an average improvement of 123.94\% and 245.45\% respectively, over the best baseline result in the PM task, and surpasses the best baseline results across three stocks in the AT task. Such an outstanding ability to balance return and risk is particularly valuable in real-world investment, where achieving high returns while effectively managing risk is crucial for long-term financial success.

\begin{table}[t]
\caption{Factor quality evaluation task results on RankIC and RankICIR (mean ± range, computed across 10 runs).}
\renewcommand{\arraystretch}{0.5}
\setlength{\abovecaptionskip}{0.1cm}
\centering
\begin{threeparttable} 
\small
\setlength{\tabcolsep}{0.5mm}{
\resizebox{0.85\linewidth}{!}{
\begin{tabular}{lccclcc} 
\toprule
 \multirow{3}{*}{\textbf{Models}} && \multicolumn{2}{c}{\textbf{SP500}} && \multicolumn{2}{c}{\textbf{DJ30}} \\ 
\cmidrule{3-4}\cmidrule{6-7}
&& RankIC$\uparrow$  & RankICIR$\uparrow$  && RankIC$\uparrow$  & RankICIR$\uparrow$   \\ 
\midrule
\multicolumn{7}{c}{\textit{Future Return Prediction in Portfolio Management Task}} \\ 
\midrule
\multirow{2}{*}{LightGBM} 	&&	0.027 	&	0.274 	&&	0.031 	&	0.272 	\\
&& \scriptsize	± 0.006	&\scriptsize	 ± 0.084	&&\scriptsize	 ± 0.005	&\scriptsize ± 0.049	\\

\multirow{2}{*}{LSTM} 	&&	0.034 	&	0.333 	&&	0.031 	&	0.329\\
&& \scriptsize	± 0.006	&\scriptsize	 ± 0.042	&&\scriptsize	 ± 0.004	&\scriptsize	 ± 0.056	\\

\multirow{2}{*}{Transformer} 	&&	0.035 	&	0.340 	&&	0.033 	&	0.343 	\\
&& \scriptsize	± 0.007	&\scriptsize	 ± 0.078	&&\scriptsize	 ± 0.005	&\scriptsize	 ± 0.045	\\

\multirow{2}{*}{CAFactor} 	&& 	0.037	&	0.356	&&	0.040	&	0.380	\\
&& \scriptsize	± 0.005	&\scriptsize	 ± 0.084	&&\scriptsize	 ± 0.003	&\scriptsize	± 0.043		\\
\multirow{2}{*}{FactorVAE} 	&& 0.052	&	0.543	&&	0.056	&	0.520		\\
&& \scriptsize	± 0.010	&\scriptsize	 ± 0.122	&&\scriptsize	 ± 0.012	&\scriptsize	± 0.081		\\
\multirow{2}{*}{HireVAE} 	&& 	\underline{0.057}	&	\underline{0.558}	&&	\underline{0.058}	&	\underline{0.563}		\\
&& \scriptsize	± 0.006	&\scriptsize	 ± 0.058	&&\scriptsize	 ± 0.006	&\scriptsize	 ± 0.053		\\

\midrule

\multirow{2}{*}{\textbf{STORM}} 	&&	\textbf{0.062}	&	\textbf{0.673}	&&	\textbf{0.065}	&	\textbf{0.668}	\\
&&\scriptsize	± 0.018	&\scriptsize	± 0.155	&&\scriptsize	± 0.038	&\scriptsize	± 0.287
\\

\multirow{2}{*}{STORM-w/o-TS} 	&& 0.053 	&	0.513 	&&	0.055 	&	0.563 	\\
&&\scriptsize	± 0.017	&\scriptsize ± 0.145	&&\scriptsize	± 0.015	&\scriptsize	± 0.127
\\

\multirow{2}{*}{STORM-w/o-CS} 	&&	0.054 	&	0.522 	&&	0.053	&	0.559 	\\
&&\scriptsize	± 0.014	&\scriptsize	± 0.118	&&\scriptsize	± 0.016	&\scriptsize	± 0.156
\\

\midrule
\multicolumn{2}{c}{Improvement(\%)\footnotemark[1]} &	8.772	&	20.609	&&	12.069	&	18.650	\\
\bottomrule
\end{tabular}
}

\begin{tablenotes}
\scriptsize	
\item[$1$]  Improvement of \texttt{STORM} over the best-performing baselines.
\end{tablenotes}
}
\end{threeparttable}
\label{table:prediction}
\end{table}

\subsection{Factor Quality (for RQ2)}
To demonstrate the effectiveness and quality of the factors, we designed a stock future return prediction task. The goal is not to pursue time-series forecasting accuracy, but to evaluate the predictive quality and robustness of the factors in capturing dynamic market patterns. In this task, \texttt{STORM} shows substantial improvements over six methods across the SP500 and DJ30 datasets. We use rank-based correlation metrics (i.e., RankIC, RankICIR) instead of error-based ones (e.g., MSE, MAE), as they better capture a factor’s ability to preserve the correct cross-sectional ordering of returns. As shown in Table ~\ref{table:prediction}, \texttt{STORM} achieves substantial average improvements of 14.690\% on SP500 and 15.360\% on DJ30, demonstrating its ability to capture more consistent and generalizable signals across diverse market conditions.
\begin{figure}[htbp]
    \setlength{\abovecaptionskip}{0.03cm}
    \centering
    \includegraphics[width=0.36\textwidth]{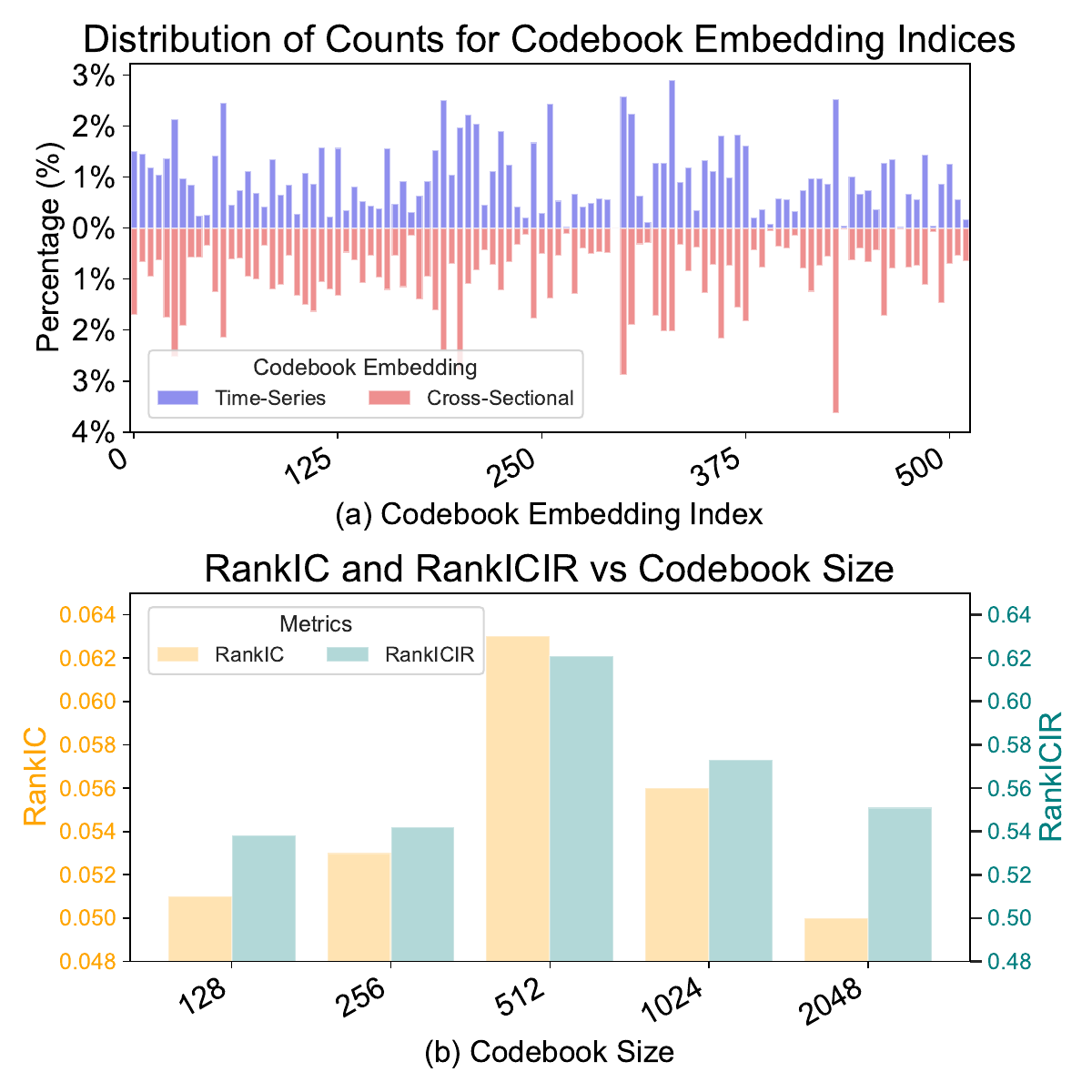}
    \caption{(a) Distribution of counts for codebook embedding indices. (b) Hyperparameter experiment results for different codebook sizes.}
    \label{fig: count}
\end{figure}

\noindent\textbf{Factor Diversity}.
The diversity of factors is essential in financial investment, as it allows the model to capture the multi-dimensional characteristics of the market, avoiding redundancy or over-reliance on a single factor. In the codebook, each embedding represents a cluster center for a category of factor embeddings. To analyze the utilization rate of each factor category, we report the frequency of codebook embedding indices in the testing data. For clarity, we group every five indices into one interval, as illustrated in Figure \ref{fig: count}(a). The results indicates a relatively even distribution of factor usage, highlighting the model’s effectiveness to maintain diversity during the factor selection process.

\noindent\textbf{Codebook Design}.
Furthermore, we conduct a hyperparameter analysis on the DJ30 dataset to examine the effect of different codebook sizes, aiming to balance factor diversity and representational quality. An excessively large codebook may lead to sparse factor representations, limiting the effectiveness of factor utilization, while an overly small codebook could result in insufficient factor differentiation and information redundancy. Following standard practices, we evaluate five different codebook sizes, including 128, 256, 512, 1024, and 2048. As shown in Figure \ref{fig: count}(b), a size of 512 achieves the best overall factor prediction performance, offering a favorable trade-off between expressiveness and efficiency.

\subsection{Ablation Study (for RQ3)}
We conducted ablation experiments with two simplified versions of \texttt{STORM}, namely \texttt{STORM}-w/o-TS and \texttt{STORM}-w/o-CS, which only extract cross-sectional and time-series factors, respectively. Specifically, the absence of either time-series or cross-sectional factors significantly reduces the model's effectiveness across various tasks. In the \textit{stock return prediction} task, \texttt{STORM} outperforms both simplified models, with an average improvement of 16.50\% on the RankIC metric and 23.79\% on the RankICIR metric, highlighting the enhanced accuracy achieved by incorporating both time-series and cross-sectional factors. In the \textit{PM} and \textit{AT} tasks, \texttt{STORM} and two simplified models all performed well. Notably, \texttt{STORM}-w/o-TS exhibited the lowest AVO among the three models, although its profitability did not surpass that of \texttt{STORM}. These findings suggest that methods like FactorVAE \cite{duan2022factorvae}, which focus solely on cross-sectional factors, may struggle to capture the full complexity of the financial market. Therefore, integrating both time-series and cross-sectional factors is essential for robust and accurate financial modeling.

\subsection{Discussion on Performance}
\noindent\textbf{Efficiency}. \texttt{STORM} runs efficiently in real-world daily trading without the ultra-low latency or specialized hardware needed for high-frequency systems, even on standard CPUs. This is enabled by a compact architecture with only a few transformer layers and 15.2M FP32 parameters on DJ30, enabling deployment on edge devices while supporting concurrent market analysis and trading.

\noindent\textbf{Effectiveness}. \texttt{STORM} delivers consistently strong performance across diverse tasks, including PM and AT, outperforming baselines that often specialize in one domain but fail to generalize. It balances return and risk, targeting comprehensive investment performance rather than over-optimizing for a single metric. Robustness and versatility across tasks justify the design of our proposed model.

\noindent\textbf{Generalization}. \texttt{STORM} generalizes across markets of varying scales and characteristics, excelling on both DJ30 (large-cap industrial blue chips) and SP500 (broad, diversified industries). These representative U.S. datasets cover concentrated and diversified indices, and their depth makes them a strong proxy for other markets like China or the U.K.. The flexible encoder-decoder architecture also allows using additional layers like Mamba ~\cite{gu2023mamba, mamba2} and ensures applicability to diverse trading environments and asset classes.

\section{Conclusions and Future Work}
In this paper, we propose \texttt{STORM}, a spatio-temporal latent factor model leveraging a dual VQ-VAE framework to learn both time-series and cross-sectional factors. \texttt{STORM} ensures factor orthogonality and diversity, crucial for effective factor selection in financial trading. 
Unlike traditional factors, \texttt{STORM} learns high-dimensional latent vectors that, though less interpretable, capture richer market structures.
Extensive experiments demonstrate \texttt{STORM}'s superiority over the state-of-the-art methods across a prediction task and two downstream tasks, showcasing robust performance and effectiveness. In the future, we plan to extend \texttt{STORM} by incorporating side information like news, to explore the effects of exogenous factors.

\section*{Acknowledgments}
This research is supported by the RIE2025 Industry Alignment Fund – Industry Collaboration Projects (IAF-ICP) (Award I2301E0026), administered by A*STAR, as well as supported by Alibaba Group and NTU Singapore through Alibaba-NTU Global e-Sustainability CorpLab (ANGEL). It is also supported, in part, by Jinan-NTU Green Technology Research Institute (GreenTRI).

\vfill\eject

\bibliographystyle{ACM-Reference-Format}
\balance
\bibliography{main}

@article{gu2021autoencoder,
  title={Autoencoder asset pricing models},
  author={Gu, Shihao and Kelly, Bryan and Xiu, Dacheng},
  journal={Journal of Econometrics},
  volume={222},
  number={1},
  pages={429--450},
  year={2021},
  publisher={Elsevier}
}

@inproceedings{duan2022factorvae,
  title={Factorvae: A probabilistic dynamic factor model based on variational autoencoder for predicting cross-sectional stock returns},
  author={Duan, Yitong and Wang, Lei and Zhang, Qizhong and Li, Jian},
  booktitle={Proceedings of the AAAI Conference on Artificial Intelligence},
  volume={36},
  number={4},
  pages={4468--4476},
  year={2022}
}

@article{wei2023hirevae,
  title={HireVAE: An Online and Adaptive Factor Model Based on Hierarchical and Regime-Switch VAE},
  author={Wei, Zikai and Rao, Anyi and Dai, Bo and Lin, Dahua},
  journal={arXiv preprint arXiv:2306.02848},
  year={2023}
}

@article{wei2022factor,
  title={Factor investing with a deep multi-factor model},
  author={Wei, Zikai and Dai, Bo and Lin, Dahua},
  journal={arXiv preprint arXiv:2210.12462},
  year={2022}
}

@article{sharpe1964capital,
  title={Capital asset prices: A theory of market equilibrium under conditions of risk},
  author={Sharpe, William F},
  journal={The Journal of Finance},
  volume={19},
  number={3},
  pages={425--442},
  year={1964},
  publisher={Wiley Online Library}
}

@article{fama1992cross,
  title={The cross-section of expected stock returns},
  author={Fama, Eugene F and French, Kenneth R},
  journal={The Journal of Finance},
  volume={47},
  number={2},
  pages={427--465},
  year={1992},
  publisher={Wiley Online Library}
}

@article{van2017neural,
  title={Neural discrete representation learning},
  author={Van Den Oord, Aaron and Vinyals, Oriol and others},
  journal={Advances in Neural Information Processing Systems},
  volume={30},
  year={2017}
}

@article{kingma2013auto,
  title={Auto-encoding variational bayes},
  author={Kingma, Diederik P and Welling, Max},
  journal={arXiv preprint arXiv:1312.6114},
  year={2013}
}

@inproceedings{zhang2024reinforcement,
  title={Reinforcement Learning with Maskable Stock Representation for Portfolio Management in Customizable Stock Pools},
  author={Zhang, Wentao and Zhao, Yilei and Sun, Shuo and Ying, Jie and Xie, Yonggang and Song, Zitao and Wang, Xinrun and An, Bo},
  booktitle={Proceedings of the ACM on Web Conference 2024},
  pages={187--198},
  year={2024}
}

@inproceedings{wang2021deeptrader,
  title={DeepTrader: a deep reinforcement learning approach for risk-return balanced portfolio management with market conditions Embedding},
  author={Wang, Zhicheng and Huang, Biwei and Tu, Shikui and Zhang, Kun and Xu, Lei},
  booktitle={Proceedings of the AAAI Conference on Artificial Intelligence},
  volume={35},
  number={1},
  pages={643--650},
  year={2021}
}

@inproceedings{ye2020reinforcement,
  title={Reinforcement-learning based portfolio management with augmented asset movement prediction states},
  author={Ye, Yunan and Pei, Hengzhi and Wang, Boxin and Chen, Pin-Yu and Zhu, Yada and Xiao, Ju and Li, Bo},
  booktitle={Proceedings of the AAAI Conference on Artificial Intelligence},
  volume={34},
  number={01},
  pages={1112--1119},
  year={2020}
}

@inproceedings{sun2022deepscalper,
  title={DeepScalper: A risk-aware reinforcement learning framework to capture fleeting intraday trading opportunities},
  author={Sun, Shuo and Xue, Wanqi and Wang, Rundong and He, Xu and Zhu, Junlei and Li, Jian and An, Bo},
  booktitle={Proceedings of the 31st ACM International Conference on Information \& Knowledge Management},
  pages={1858--1867},
  year={2022}
}

@inproceedings{liu2020adaptive,
  title={Adaptive quantitative trading: An imitative deep reinforcement learning approach},
  author={Liu, Yang and Liu, Qi and Zhao, Hongke and Pan, Zhen and Liu, Chuanren},
  booktitle={Proceedings of the AAAI Conference on Artificial Intelligence},
  volume={34},
  number={02},
  pages={2128--2135},
  year={2020}
}

@article{ross1976arbitrage,
  title={The arbitrage pricing theory},
  author={Ross, Stephen},
  journal={Journal of Economic Theory},
  volume={13},
  number={3},
  pages={341--360},
  year={1976}
}

@article{vaswani2017attention,
  title={Attention is all you need},
  author={Vaswani, Ashish and Shazeer, Noam and Parmar, Niki and Uszkoreit, Jakob and Jones, Llion and Gomez, Aidan N and Kaiser, {\L}ukasz and Polosukhin, Illia},
  journal={Advances in Neural Information Processing Systems},
  volume={30},
  year={2017}
}

@article{nie2022time,
  title={A time series is worth 64 words: Long-term forecasting with transformers},
  author={Nie, Yuqi and Nguyen, Nam H and Sinthong, Phanwadee and Kalagnanam, Jayant},
  journal={arXiv preprint arXiv:2211.14730},
  year={2022}
}

@inproceedings{chen2023mtrader,
  title={mTrader: A Multi-Scale Signal Optimization Deep Reinforcement Learning Framework for Financial Trading (S).},
  author={Chen, Zhennan and Zhang, Zhicheng and Li, Pengfei and Wei, Lingyue and Feng, Shibo and Lin, Fan},
  booktitle={SEKE},
  pages={530--535},
  year={2023}
}

@article{zhang2024finagent,
  title={FinAgent: A Multimodal Foundation Agent for Financial Trading: Tool-Augmented, Diversified, and Generalist},
  author={Zhang, Wentao and Zhao, Lingxuan and Xia, Haochong and Sun, Shuo and Sun, Jiaze and Qin, Molei and Li, Xinyi and Zhao, Yuqing and Zhao, Yilei and Cai, Xinyu and others},
  journal={arXiv preprint arXiv:2402.18485},
  year={2024}
}

@article{schulman2017proximal,
  title={Proximal policy optimization algorithms},
  author={Schulman, John and Wolski, Filip and Dhariwal, Prafulla and Radford, Alec and Klimov, Oleg},
  journal={arXiv preprint arXiv:1707.06347},
  year={2017}
}

@inproceedings{shin2023exploration,
  title={Exploration into translation-equivariant image quantization},
  author={Shin, Woncheol and Lee, Gyubok and Lee, Jiyoung and Lyou, Eunyi and Lee, Joonseok and Choi, Edward},
  booktitle={ICASSP 2023-2023 IEEE International Conference on Acoustics, Speech and Signal Processing (ICASSP)},
  pages={1--5},
  year={2023},
  organization={IEEE}
}

@article{baevski2020wav2vec,
  title={wav2vec 2.0: A framework for self-supervised learning of speech representations},
  author={Baevski, Alexei and Zhou, Yuhao and Mohamed, Abdelrahman and Auli, Michael},
  journal={Advances in Neural Information Processing Systems},
  volume={33},
  pages={12449--12460},
  year={2020}
}

@inproceedings{chen2021crossvit,
  title={Crossvit: Cross-attention multi-scale vision transformer for image classification},
  author={Chen, Chun-Fu Richard and Fan, Quanfu and Panda, Rameswar},
  booktitle={Proceedings of the IEEE/CVF international Conference on Computer Vision},
  pages={357--366},
  year={2021}
}

@inproceedings{li2023blip,
  title={Blip-2: Bootstrapping language-image pre-training with frozen image encoders and large language models},
  author={Li, Junnan and Li, Dongxu and Savarese, Silvio and Hoi, Steven},
  booktitle={International Conference on Machine Learning},
  pages={19730--19742},
  year={2023},
  organization={PMLR}
}

@inproceedings{radford2021learning,
  title={Learning transferable visual models from natural language supervision},
  author={Radford, Alec and Kim, Jong Wook and Hallacy, Chris and Ramesh, Aditya and Goh, Gabriel and Agarwal, Sandhini and Sastry, Girish and Askell, Amanda and Mishkin, Pamela and Clark, Jack and others},
  booktitle={International Conference on Machine Learning},
  pages={8748--8763},
  year={2021},
  organization={PMLR}
}

@article{dosovitskiy2020image,
  title={An image is worth 16x16 words: Transformers for image recognition at scale},
  author={Dosovitskiy, Alexey and Beyer, Lucas and Kolesnikov, Alexander and Weissenborn, Dirk and Zhai, Xiaohua and Unterthiner, Thomas and Dehghani, Mostafa and Minderer, Matthias and Heigold, Georg and Gelly, Sylvain and others},
  journal={arXiv preprint arXiv:2010.11929},
  year={2020}
}

@book{greenblatt2010little,
  title={The little book that still beats the market},
  author={Greenblatt, Joel},
  volume={29},
  year={2010},
  publisher={John Wiley \& Sons}
}

@article{sun2024trademaster,
  title={TradeMaster: a holistic quantitative trading platform empowered by reinforcement learning},
  author={Sun, Shuo and Qin, Molei and Zhang, Wentao and Xia, Haochong and Zong, Chuqiao and Ying, Jie and Xie, Yonggang and Zhao, Lingxuan and Wang, Xinrun and An, Bo},
  journal={Advances in Neural Information Processing Systems},
  volume={36},
  year={2024}
}

@article{li2014online,
  title={Online portfolio selection: A survey},
  author={Li, Bin and Hoi, Steven CH},
  journal={ACM Computing Surveys (CSUR)},
  volume={46},
  number={3},
  pages={1--36},
  year={2014},
  publisher={ACM New York, NY, USA}
}

@article{poterba1988mean,
  title={Mean reversion in stock prices: Evidence and implications},
  author={Poterba, James M and Summers, Lawrence H},
  journal={Journal of Jinancial Economics},
  volume={22},
  number={1},
  pages={27--59},
  year={1988},
  publisher={Elsevier}
}

@article{ke2017lightgbm,
  title={Lightgbm: A highly efficient gradient boosting decision tree},
  author={Ke, Guolin and Meng, Qi and Finley, Thomas and Wang, Taifeng and Chen, Wei and Ma, Weidong and Ye, Qiwei and Liu, Tie-Yan},
  journal={Advances in Neural Information Processing Systems},
  volume={30},
  year={2017}
}

@article{qin2017dual,
  title={A dual-stage attention-based recurrent neural network for time series prediction},
  author={Qin, Yao and Song, Dongjin and Chen, Haifeng and Cheng, Wei and Jiang, Guofei and Cottrell, Garrison},
  journal={arXiv preprint arXiv:1704.02971},
  year={2017}
}

@article{zhao2022stock,
  title={Stock movement prediction based on bi-typed hybrid-relational market knowledge graph via dual attention networks},
  author={Zhao, Yu and Du, Huaming and Liu, Ying and Wei, Shaopeng and Chen, Xingyan and Zhuang, Fuzhen and Li, Qing and Kou, Gang},
  journal={IEEE Transactions on Knowledge and Data Engineering},
  volume={35},
  number={8},
  pages={8559--8571},
  year={2022},
  publisher={IEEE}
}

@inproceedings{fang2021universal,
  title={Universal trading for order execution with oracle policy distillation},
  author={Fang, Yuchen and Ren, Kan and Liu, Weiqing and Zhou, Dong and Zhang, Weinan and Bian, Jiang and Yu, Yong and Liu, Tie-Yan},
  booktitle={Proceedings of the AAAI Conference on Artificial Intelligence},
  volume={35},
  number={1},
  pages={107--115},
  year={2021}
}

@article{spooner2020robust,
  title={Robust market making via adversarial reinforcement learning},
  author={Spooner, Thomas and Savani, Rahul},
  journal={arXiv preprint arXiv:2003.01820},
  year={2020}
}

@article{yang2020qlib,
  title={Qlib: An ai-oriented quantitative investment platform},
  author={Yang, Xiao and Liu, Weiqing and Zhou, Dong and Bian, Jiang and Liu, Tie-Yan},
  journal={arXiv preprint arXiv:2009.11189},
  year={2020}
}

@article{talukder2024totem,
  title={TOTEM: TOkenized Time Series EMbeddings for General Time Series Analysis},
  author={Talukder, Sabera and Yue, Yisong and Gkioxari, Georgia},
  journal={arXiv preprint arXiv:2402.16412},
  year={2024}
}

@article{yan2021videogpt,
  title={Videogpt: Video generation using vq-vae and transformers},
  author={Yan, Wilson and Zhang, Yunzhi and Abbeel, Pieter and Srinivas, Aravind},
  journal={arXiv preprint arXiv:2104.10157},
  year={2021}
}

@inproceedings{zhang2024codebook,
  title={Codebook Transfer with Part-of-Speech for Vector-Quantized Image Modeling},
  author={Zhang, Baoquan and Wang, Huaibin and Luo, Chuyao and Li, Xutao and Liang, Guotao and Ye, Yunming and Qi, Xiaochen and He, Yao},
  booktitle={Proceedings of the IEEE/CVF Conference on Computer Vision and Pattern Recognition},
  pages={7757--7766},
  year={2024}
}

@article{razavi2019generating,
  title={Generating diverse high-fidelity images with vq-vae-2},
  author={Razavi, Ali and Van den Oord, Aaron and Vinyals, Oriol},
  journal={Advances in Neural Information Processing Systems},
  volume={32},
  year={2019}
}

@inproceedings{yu2018forecasting,
  title={Forecasting stock price index volatility with LSTM deep neural network},
  author={Yu, ShuiLing and Li, Zhe},
  booktitle={Recent Developments in Data Science and Business Analytics: Proceedings of the International Conference on Data Science and Business Analytics (ICDSBA-2017)},
  pages={265--272},
  year={2018},
  organization={Springer}
}

@article{mnih2013playing,
  title={Playing atari with deep reinforcement learning},
  author={Mnih, Volodymyr and Kavukcuoglu, Koray and Silver, David and Graves, Alex and Antonoglou, Ioannis and Wierstra, Daan and Riedmiller, Martin},
  journal={arXiv preprint arXiv:1312.5602},
  year={2013}
}

@article{haarnoja2018soft,
  title={Soft actor-critic algorithms and applications},
  author={Haarnoja, Tuomas and Zhou, Aurick and Hartikainen, Kristian and Tucker, George and Ha, Sehoon and Tan, Jie and Kumar, Vikash and Zhu, Henry and Gupta, Abhishek and Abbeel, Pieter and others},
  journal={arXiv preprint arXiv:1812.05905},
  year={2018}
}

@article{wang2021clvsa,
  title={CLVSA: A convolutional LSTM based variational sequence-to-sequence model with attention for predicting trends of financial markets},
  author={Wang, Jia and Sun, Tong and Liu, Benyuan and Cao, Yu and Zhu, Hongwei},
  journal={arXiv preprint arXiv:2104.04041},
  year={2021}
}

@article{yu2015multi,
  title={Multi-scale context aggregation by dilated convolutions},
  author={Yu, Fisher and Koltun, Vladlen},
  journal={arXiv preprint arXiv:1511.07122},
  year={2015},
  pages={1--10}
}

@article{gu2023mamba,
  title={Mamba: Linear-time sequence modeling with selective state spaces},
  author={Gu, Albert and Dao, Tri},
  journal={arXiv preprint arXiv:2312.00752},
  year={2023}
}

@inproceedings{mamba2,
  title={Transformers are {SSM}s: Generalized Models and Efficient Algorithms Through Structured State Space Duality},
  author={Dao, Tri and Gu, Albert},
  booktitle={International Conference on Machine Learning (ICML)},
  year={2024}
}

\end{document}